\def\year{2018}\relax
\documentclass[letterpaper]{article} 
\usepackage{aaai18}  
\usepackage{times}  
\usepackage{helvet}  
\usepackage{courier}  
\usepackage{url}  
\usepackage{graphicx}  
\usepackage{nameref}
\usepackage[utf8]{inputenc} 
\usepackage[T1]{fontenc}    
\usepackage{url}            
\usepackage{booktabs}       
\usepackage{amsfonts}       
\usepackage{nicefrac}       
\usepackage{microtype}      
\usepackage{wrapfig}
\usepackage{floatflt} 
\usepackage{paralist} 
\usepackage{algorithm} 
\usepackage{amsmath}
\usepackage{amsthm}
\usepackage{amssymb}
\usepackage{footmisc}
\usepackage{listings}  
\usepackage{xcolor}
\usepackage{color}
\usepackage{colortbl} 
\usepackage{indentfirst}
\usepackage{calligra}
\usepackage{bm}
\usepackage{fancyhdr}
\usepackage{pifont}
\usepackage{algpseudocode} 
\usepackage[export]{adjustbox}

\newcommand{\mypar}[1]{{\bf #1.}}
\theoremstyle{definition}
\newtheorem{defn}{Definition}

\newtheorem{myThm}{Theorem}

\def\x{\mathbf{x}}
\def\g{\mathbf{g}}
\def\r{\mathbf{r}}

\def\y{\mathbf{y}}

\def\q{\mathbf{q}}

\def\t{\mathbf{t}}
\def\vv{\mathbf{v}}
\def\w{\mathbf{w}}

\def\V{\mathcal{V}}
\def\E{\mathcal{E}}

\DeclareMathOperator{\Adj}{\mathbf{A}}

\DeclareMathOperator{\RR}{\mathbf{R}}
\DeclareMathOperator{\Pj}{\mathbf{P}}

\DeclareMathOperator{\X}{\mathbf{X}}

\newcommand{\R}{\ensuremath{\mathbb{R}}}

\DeclareMathOperator{\Id}{I}

\begin{document}
%
\title{Generalized Value Iteration Networks: \\ Life Beyond Lattices}
\author{
  Sufeng Niu\thanks{Equal contribution.}\footnotemark[2] Siheng Chen\footnotemark[1]\footnotemark[3]
  , Hanyu Guo\footnotemark[1]\footnotemark[2], Colin Targonski\footnotemark[2], Melissa C. Smith\footnotemark[2], Jelena Kovačević\footnotemark[3] \\
  \footnotemark[2] Clemson University, 433 Calhoun Dr., Clemson, SC 29634, USA \\
  \footnotemark[3] Carnegie Mellon University, 5000 Forbes Avenue, Pittsburgh, PA 15213, USA \\
}
\maketitle
\begin{abstract}
  In this paper, we introduce a generalized value iteration network (GVIN), which is an end-to-end neural network planning module. GVIN emulates the value iteration algorithm by using a novel graph convolution operator, which enables GVIN to learn and plan on irregular spatial graphs. We propose three novel differentiable kernels as graph convolution operators and show that the embedding-based kernel achieves the best performance. Furthermore, we present episodic $Q$-learning, an improvement upon traditional $n$-step $Q$-learning that stabilizes training for VIN and GVIN.
  Lastly, we evaluate GVIN on planning problems in 2D mazes, irregular graphs, and real-world street networks, showing that GVIN generalizes well for both arbitrary graphs and unseen graphs of larger scale
and outperforms a naive generalization of VIN  (discretizing a spatial graph into a 2D image). 
\end{abstract}

\vspace{-5mm}
\section{Introduction}
\label{sec:intro}
\vspace{-1mm}

Reinforcement learning (RL) is a technique that solves sequential decision making problems that lacks explicit rules and labels~\cite{sutton1998reinforcement}. Recent developments in Deep Reinforcement Learning(DRL) have lead to enormous progress in autonomous driving~\cite{bojarski2016end}, innovation in robot control~\cite{levine2015end}, and human-level performance in both Atari games~\cite{mnih2013playing,guo2014deep} and the board game Go~\cite{silver2016mastering}. Given a reinforcement learning task, the agent explores the underlying Markov Decision Process (MDP)~\cite{bellman1957dynamic,bertsekas1995dynamic} and attempts to learn a mapping of high-dimensional state space data to an optimal policy that maximizes the expected return. Reinforcement learning can be categorized as model-free~\cite{lillicrap2015continuous,mnih2016asynchronous,mnih2013playing} and model-based approaches~\cite{sutton1998reinforcement,deisenroth2011pilco,schmidhuber1990line}. Model-free approaches learn the policy directly by trial-and-error and attempt to avoid bias caused by a suboptimal environment model~\cite{sutton1998reinforcement}; the majority of recent architectures for DRL follow the model-free approach~\cite{lillicrap2015continuous,mnih2016asynchronous,mnih2013playing}. Model-based approaches, on the other hand, allow for an agent to explicitly learn the mechanisms of an environment, which can lead to strong generalization abilities. A recent work, the value iteration networks (VIN)~\cite{tamar2016value} combines recurrent convolutional neural networks and max-pooling to emulate the process of value iteration~\cite{bellman1957dynamic,bertsekas1995dynamic}. As VIN learns an environment, it can plan shortest paths for unseen mazes.

The input data fed into deep learning systems is usually associated with regular structures. For example, speech signals and natural language have an underlying 1D sequential structure; images have an underlying 2D lattice structure. To take advantage of this regularly structured data, deep learning uses a series of basic operations defined for the regular domain, such as convolution and uniform pooling. However, not all data is contained in regular structures. In urban science, traffic information is associated with road networks; in neuroscience, brain activity is associated with brain connectivity networks; in social sciences, users' profile information is associated with social networks. To learn from data with irregular structure, some recent works have extended the lattice structure to general graphs~\cite{defferrard2016convolutional,kipf2016semi} and redefined convolution and pooling operations on graphs; however, most works only evaluate data that has both a fixed and given graph. In addition, most lack the ability to generalize to new, unseen environments.


In this paper, we aim to enable an agent to self-learn and plan the optimal path in new, unseen spatial graphs by using model-based DRL and graph-based techniques. 
This task is relevant to many real-world applications, such as route planning of self-driving cars and web crawling/navigation. The proposed method is  more general than classical DRL, extending for irregular structures. 
Furthermore, the proposed method is scalable (computational complexity is proportional to the number of edges in the testing graph), handles various edge weight settings and adaptively learns the environment model. Note that the optimal path can be self-defined, and is not necessarily the shortest one. Additionally, the proposed work differs from conventional planning algorithms; for example, Dijkstra's algorithm requires a known model, while GVIN aims to learn a general model via trial and error, then apply said model to new, unseen irregular graphs.

To create GVIN, we generalize VIN in two aspects. First, to work for irregular graphs, we propose a graph convolution operator that generalizes the original 2D convolution operator.  With the new graph convolution operator, the proposed network captures the basic concepts in spatial graphs, such as direction, distance and edge weight. It also is able to transfer knowledge learned from one graph to others. Second, to improve reinforcement learning on irregular graphs,  we propose a reinforcement learning algorithm, episodic $Q$-learning, which stabilizes the training for VIN and GVIN. The original VIN is trained through either imitation learning, which requires a large number of ground-truth labels, or reinforcement learning, whose performance is relatively poor. With the proposed episodic $Q$-learning, the new network performs significantly better than VIN in the reinforcement learning mode.  Since the proposed network generalizes the original VIN model, we call it the~\emph{generalized value iteration network (GVIN)}. 


The main contributions of this paper are: \\
$\bullet$ The proposed architecture, GVIN, generalizes the VIN ~\cite{tamar2016value} to handle both regular structures and irregular structures. GVIN offers an end-to-end architecture trained via reinforcement learning (no ground-truth labels); see Section~\textit{\nameref{sec:framework}}; \\
$\bullet$ The proposed graph convolution operator generalizes 2D convolution learns the concepts of direction and distance, which enables GVIN to transfer knowledge from one graph to another; see Section~\textit{\nameref{sec:graph_convolution}}; \\
$\bullet$ The proposed reinforcement learning algorithm, episodic $Q$-learning, extends the classical $n$-step $Q$-learning as Monte Carlo control and significantly improves the performance of reinforcement learning for irregular graphs; see Section~\textit{\nameref{sec:train_rl}}; and \\
$\bullet$ Through intensive experiments we demonstrate the generalization ability of GVIN within imitation learning and episodic $Q$-learning for various datasets, including synthetic 2D maze data, irregular graphs, and real-world maps (Minnesota highway and New York street maps); we show that GVIN significantly outperforms VIN with discretization input on irregular structures; See Section~\textit{\nameref{sec:exp}}.

\vspace{-3mm}
\section{Background}
\vspace{-1mm}

\mypar{Markov Decision Process}
We consider an environment defined as an MDP that contains a set of states $s\in S$, a set of actions $a\in A$,  a reward function $\RR_{s,a}$, and a series of transition probabilities $\Pj_{s',s,a}$, the probability of moving from the current state $s$ to  the next state $s'$ given an action $a$.  The goal of an MDP is to find a policy that maximizes the expected return (accumulated rewards) $R_t=\sum_{k=0}^\infty\gamma^k r_{t+k}$, where $r_{t+k}$ is the immediate reward at the $(t+k)$th time stamp and $\gamma\in (0,1]$ is the discount rate.  A policy $\pi_{a,s}$ is the probability of taking action $a$ when in state $s$.  The value of state $s$ under a policy $\pi$, $\vv^{\pi}_{s}$, is the expected return when starting in $s$ and following $\pi$; that is, $\vv^{\pi}_{s} = \mathbb{E}[R_t | S_t = s]$.  The value of taking action $a$ in state $s$ under a policy $\pi$,  $ {\q^{\pi}}_s^{(a)}$, is the expected return when starting in $s$, taking the action $a$ and following $\pi$; that is, $ {\q^{\pi}}_s^{(a)} = \mathbb{E}[R_t | S_t = s, A_t = a]$.  There is at least one policy that is better than or equal to all other policies, called an optimal policy $\pi^*$; that is, the  optimal policy is $\pi^* = \arg \max_{\pi} \vv^{\pi}_{s}$, the optimal state-value function is  $\vv^{*}_{s} = \max_{\pi} \vv^{\pi}_{s}$, and the optimal action-value function is  $ {\q^*}_s^{(a)}  = \max_{\pi} {\q^{\pi}}_s^{(a)}$.  To obtain $\pi^*$ and $\vv^*$, we usually consider solving the Bellman equation. Value iteration is a popular algorithm used to solve the Bellman equation in the discrete state space; that is, we iteratively compute $\vv_{s} \leftarrow  \max_a\sum_{s'} \Pj_{s',s,a} \left( \RR_{s,a} +\gamma \vv_{s'} \right)$ until  convergence.

\mypar{Differentiable planning module}
VIN employs an embedded differentiable planning architecture, trained end-to-end via imitation learning~\cite{tamar2016value}. In VIN, the Bellman equation is encoded within the convolutional neural networks, and  the policy can be obtained through backpropagation. However, VIN is limited to regular lattices; it requires imitation learning for maximum performance and is trained separately with a reactive policy. A more recent work Memory Augmented Control Network (MACN)~\cite{khan2017memory} combines the VIN model with a memory augmented controller, which can then backtrack through the history of previous trajectories. However, as we shown later in Table~\ref{table:irregular}, GVIN outperform MACN on both performance and problem scales. A different model-based work, Predictron, uses a learning and planning model that simulates a Markov reward process~\cite{silver2016predictron}. The architecture unrolls the "imagined" plan via a predictron core. However, Predictron is limited to the Markov rewards process and is relatively computationally expensive compared to VIN.  


\mypar{Deep Learning with Graphs}
A number of recent works consider using neural networks to handle signals supported on graphs~\cite{niepert2016learning,duvenaud2015convolutional,henaff2015deep}.  The principal idea is to generalize basic operations in the regular domain, such as filtering and pooling, to the graph domain based on spectral graph theory. For example,~\cite{bruna2013spectral,henaff2015deep} introduce hierarchical clustering on graphs and the spectrum of the graph Laplacian to neural networks;~\cite{defferrard2016convolutional}  generalizes classical convolutional neural networks by using graph coarsening and localized convolutional graph filtering;~\cite{kipf2016semi} considers semi-supervised learning with graphs by using graph-based convolutional neural networks; ~\cite{li2015gated} investigate learning graph structure through gated recurrent unit;  ~\cite{gilmer2017neural} considers a message passing framework that unifies previous work, see some recent overviews in~\cite{bronstein2016geometric}. 



\begin{figure*}[htb]
  \begin{center}
  \includegraphics[width= 0.78\textwidth]{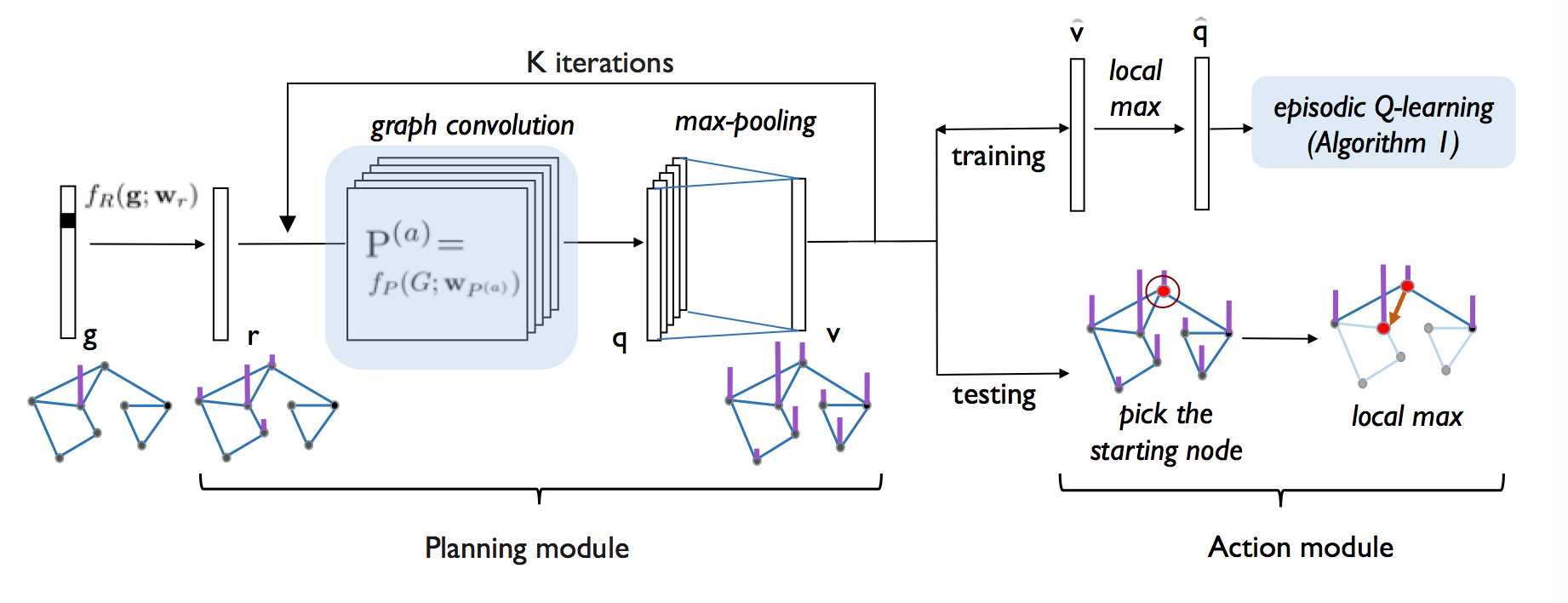}
  \vspace{-5mm}
    \caption{\label{fig:gvin} Architecture of GVIN.  The left module emulates value iteration and obtains the state values; the right module is responsible for selecting an action based on an $\epsilon$-greedy policy (for training) or a greed policy (for testing). We emphasize our contributions, including graph convolution operator and episodic $Q$-learning, in the blue blocks. }
      \vspace{-5mm}
  \end{center}
\end{figure*}

\vspace{-3mm}
\section{Methodology}
\vspace{-1mm}

We propose a new model-based DRL framework, GVIN, that takes a general graph with a starting node and a goal node as inputs and outputs the designed plan. The goal of GVIN is to learn an underlying MDP that summarizes the optimal planning policy applied for arbitrary graphs, which requires GVIN to capture general knowledge about planning that is structure and transition invariant and does not depend on any specific graph structure. A key component of an MDP is the transition matrix, which is needed to solve the Bellman equation. To train a general transition matrix that works for arbitrary graphs, similar to the VIN, we treat it as a graph convolution operator and parameterize it by using graph-based kernel functions, each of which represents a unique action pattern. We train the parameters in GVIN by using episodic $Q$-learning, which makes reinforcement learning on irregular graphs practical.

\vspace{-2mm}
\subsection{Framework}
\label{sec:framework}
\vspace{-1mm}
The input of GVIN is a graph with a starting node and a goal node. In the training phase, GVIN trains the parameters by trial-and-error on various graphs; during the testing phase, GVIN plans the optimal path based on the trained parameters. The framework includes the planning module (left) and the action module (right), shown in Figure~\ref{fig:gvin}. The planning module emulates value iteration by iteratively operating the graph convolution and max-pooling. The action module takes the greedy action according to the value function.

Mathematically, we consider a directed, weighted spatial graph $G=(\V, \X, \E, \Adj)$, where $\V=\{v_1,...,v_N\}$ is the node set, $\X \in \R^{N \times 2}$ are the node embeddings with the $i$th row $\X_i \in \R^2$  being the embedding of the $i$th node (here we consider 2D spatial graphs, but the method is generalizable), $\E=\{e_1,...,e_M\}$ is the edge set, and  $\Adj \in \mathbb{R}^{N\times N}$ is the adjacency matrix, with the $(i,j)$th element $\Adj_{i,j}$ representing the edge weight between the $i$th and $j$th nodes.  We consider a~\emph{graph signal} as a mapping from the nodes to real values. We use a graph signal $\g \in \{0, 1\}^N$ to encode the goal node, where $\g$ is one-sparse and only activates the goal node. Let $\r \in  \R^N$, $\vv  \in  \R^N$, and $\q  \in  \R^N$ be the reward graph signal, the state-value graph signal, and the action-value graph signal, respectively. We represent the entire process in a matrix-vector form as follows,
\begin{eqnarray}
\label{eq:reward}
\r & = & f_R(\g; \w_{\r}),
\\
\label{eq:graph_convolution}
\Pj^{(a)} & = & f_P( G; \w_{{\Pj}^{(a)}}),
\\
\label{eq:bellman}
\q_{n+1}^{(a)} & = & \Pj^{(a)} \left( \r + \gamma \vv_{n} \right),
\\
\label{eq:value_map}
\vv_{n+1}  & = &  \max_a  \q_{n+1}^{(a)}.
\end{eqnarray}

In the feature-extraction step~\eqref{eq:reward}, $\g$ is encoded to become the robust reward $\r$ via the feature-extract function $f_R(\cdot)$, which is a convolutional neural network in the case of regular graphs, but is the identity function when operating on irregular graphs; in step~\eqref{eq:graph_convolution}, where $\Pj^{(a)}$ is the graph convolution operator in the $a$th channel, a set of graph convolution operators is trained based on the graph $G$, which is further described in Section~\textit{\nameref{sec:graph_convolution}}; in~\eqref{eq:bellman} and~\eqref{eq:value_map}, value iteration is emulated by using graph convolution to obtain the action-value graph signal $\q^{(a)}$ in the $a$th channel and max-pooling to obtain the state-value graph signal $\vv$. $\w_{\r}$ and $\w_{\Pj^{(a)}}$ are training parameters to parameterize $\r$ and $\Pj^{(a)}$, respectively. As shown in Figure~\ref{fig:gvin}, we repeat the graph convolution operation~\eqref{eq:bellman} and max-pooling~\eqref{eq:value_map} for $K$ iterations to obtain the final state-value graph signal $\widehat{\vv}$. When $G$ is a 2D lattice, the planning module of GVIN degenerates to VIN.


In the training phase, we feed the final state-value graph signal $\widehat{\vv}$ to the action module. The original VIN extracts the action values from step~\eqref{eq:bellman} and trains the final action probabilities for eight directions; however, this is problematic for irregular graphs, as the number of actions (neighbors) at each node varies. To solve this, we consider converting $\widehat{\vv}$ to a pseudo action-value graph signal, $\widehat{\q} \in \R^N$, whose $s$th element is $\widehat{\q}_{s} = \max_{s' \in {\rm Nei}(s)} \widehat{\vv}_{s'}$, representing the action value moving from $s$ to one of its neighbors.  The advantages of this approach come from the following three aspects:  (1) the final state value of each node is obtained by using the maximum action values across all the channels, which is robust to small variations; (2) the pseudo action-value graph signal considers a unique action for each node and does not depend on the number of actions; that is, at each node, the agent queries the state values of its neighbors and always moves to the one with the highest value; and (3)  the pseudo action-value graph signal considers local graph structure, because the next state is always chosen from one of the neighbors of the current state. 

The pseudo action-value graph signal is used through episodic $Q$-learning, which learns from trial-and-error experience and backpropagates to update all of the training parameters. In episodic $Q$-learning, each episode is obtained as follows: for each given starting node $s_0$, the agent will move sequentially from $s_{t}$ to $s_{t+1}$ by the $\epsilon$-greedy strategy; that is, with probability $(1-\epsilon)$, $s_{t+1} = \arg \max_{s' \in {\rm Nei}(s_t)} \widehat{\vv}_{s'}$ and  with probability $\epsilon$,  $s_{t+1} $ is randomly selected from one of the neighbors of $s_t$. An episode terminates when $s_{t+1}$ is the goal state or the maximum step threshold is reached. For each episode, we consider the loss function as,
$
L (\w ) = \sum_{t=1}^T \left( R_t -  \widehat{\q}_{s_t} \right)^2,
$
where  $\widehat{\q}_{s_t}$ is a function of the training parameters $\w =  [\w_{\r}, \w_{\Pj^{(a)}}]$ in GVIN, 
$T$ is the episode length and  $R_t$ is the expected return at time stamp $t$, defined as $R_{t}=(r_{t+1}+\gamma R_{t+1})$, where $\gamma$ is the discount factor, and $r_t$ is the immediate return at time stamp $t$. Additional details of the algorithm will be discussed in Section~\textit{\nameref{sec:train_rl}}. In the testing phase, we obtain the action by greedily selecting the maximal state value; that is, $s_{t+1} = \arg \max_{s' \in {\rm Nei}(s_t)} \widehat{\vv}_{s'}$.


\subsection{Graph Convolution}
\label{sec:graph_convolution}
The conventional CNN takes an image as input, which is a 2D lattice graph. Each node is a pixel and has the same local structure, sitting on a grid and connecting to its eight neighbors. In this case, the convolution operator is easy to obtain. In irregular graphs, however, nodes form diverse local structures, making it challenging to obtain a structured and translation invariant operator that transfers knowledge from one graph to another. The fundamental problem here is to find a convolution operator that works for arbitrary local structures. We solve this through learning a 2D spatial kernel function that provides a transition probability distribution in the 2D space, and according to which we evaluate the weight of each edge and obtain a graph convolution operator.

The 2D spatial kernel function assigns a value to each position in the 2D space, which reflects the possibility to transit to the corresponding position. Mathematically, the transition probability from a starting position $\x \in \R^2$ to another position $\y \in \R^2$ is $K(\x, \y)$, where $K(\cdot,\cdot)$ is a 2D spatial kernel function, which will be specified later. 
\begin{defn}
A 2D spatial kernel function $K(\cdot,\cdot)$ is shift invariant when it satisfies $K(\x, \y) = K(\x + \t, \y + \t)$, for all $\x, \y, \t \in \R^2$.
\end{defn}
The shift invariance requires that the transition probability depend on the relative position, which is the key for transfer learning; in other words, no matter where the starting position is, the transition probability distribution is invariant.  Based on a shift-invariant 2D spatial kernel function  and the graph adjacency matrix,  we obtain the graph convolution operator $\Pj =  f_P( G; \w_{\Pj}) \in \R^{N \times N}$, where each element $\Pj_{i,j} = \Adj_{i,j} \cdot K_{\w_{\Pj}}(\X_i, \X_j)$, where the kernel function  $K_{\w_{\Pj}}( \cdot, \cdot)$ is parameterized by $\w_{\Pj}$ and $\X_i, \X_j \in \R^2$ are the embeddings of the $i$th and $j$th node.  The graph convolution operator follows from (1) graph connectivity and (2) 2D spatial kernel function. With the shift-invariant property, the 2D spatial kernel function leads to the same local transition distribution at each node; the graph adjacency matrix works as a modulator to select activations in the graph convolution operator.  When there is no edge between $i$ and $j$, we have $\Adj_{i,j} = 0$ and $\Pj_{i,j} = 0$; when there is an edge between $i$ and $j$, $\Pj_{i,j}$ is high when $K_{\w_{\Pj}}(\X_i, \X_j)$ is high; in other words, when the transition probability from the $i$th node to the $j$th node is higher, the edge weight $\Pj_{i,j}$ is high and the influence from the $i$th node to the $j$th node is bigger during the graph convolution. Note that $\Pj$ is a sparse matrix and its sparsity pattern is the same with its corresponding adjacency matrix, which ensures cheap computation.



\begin{figure}[h]
	\begin{center}
		\begin{tabular}{cc}
			\includegraphics[width=0.4\textwidth]{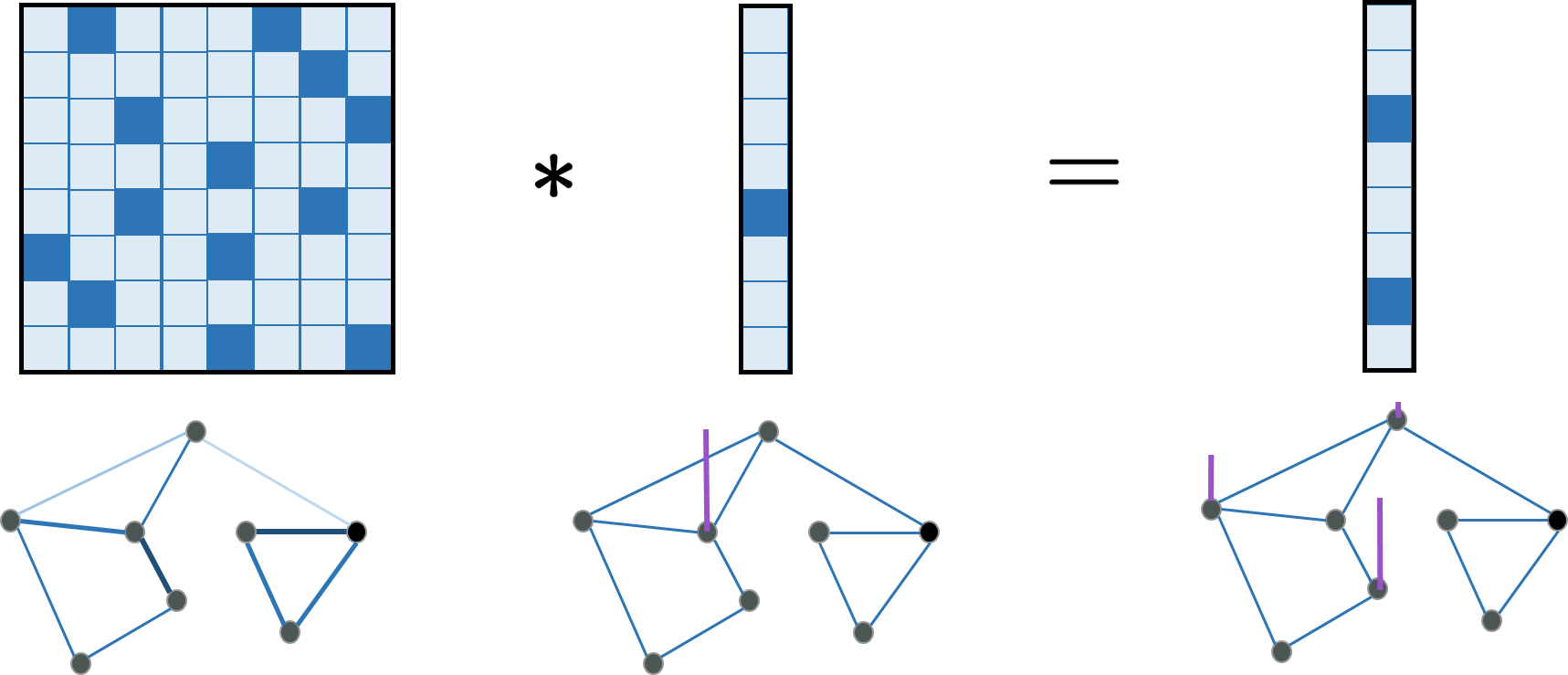}
		\end{tabular}
	\end{center}
	\vspace{-3mm}
	\caption{\label{fig:graph_convolution}  Matrix-vector multiplication as graph convolution.  Through a graph	convolution operator $\Pj$, $\r + \gamma \vv$ diffuses over the graph to obtain the action-value graph signal $\q$.
	}
	\vspace{-3mm}
\end{figure}

As shown in~\eqref{eq:graph_convolution}, the graph convolution is a matrix-vector multiplication between the graph convolution operator $\Pj$  and the graph signal $\r + \gamma \vv_{n}$; see Figure~\ref{fig:graph_convolution}. Note that when we work with a lattice graph and an appropriate kernel function, this graph convolution operator $\Pj$ is nothing but a matrix representation of the conventional convolution~\cite{lecun1995convolutional}; in other words, VIN is a special case of GVIN when the underlying graph is a 2D lattice; see more details in Supplementary~\textit{\nameref{sec:kernel}}.


We consider three types of shift-invariant 2D spatial kernel functions: the directional kernel, the spatial kernel, and the embedding kernel.
\vspace{-3mm}

\mypar{Directional Kernel} The directional kernel is embedded with the direction information.  The $(i,j)$th element in the graph convolution operator models the probability of following the edge from $i$ to $j$ ; that is, 
\vspace{-1mm}
\begin{eqnarray}
\vspace{-5mm}
\label{eq:dir_kernel}
&&\Pj_{i,j}  \ = \  \Adj_{i,j} \cdot \sum_{\ell=1}^L w_{\ell}  K^{(t,\theta_{\ell} )}_{\rm d} \left(\theta_{ij} \right), 
\\ \nonumber
&& {\rm where}~ K^{(t,\theta_{\ell} )}_{\rm d} \left(\theta \right) \ = \ \left(\frac{1+\cos(\theta - \theta_{\ell})}{2} \right)^t,
\vspace{-4mm}
\end{eqnarray}
$w_{\ell}$ is kernel coefficient,  $\theta_{ij}$ is the direction of the edge connecting the $i$th and the$j$th nodes, which can be computed through the node embeddings $\X_i,\X_j  \in \R^2$, and $ K^{(t,\theta_{\ell} )}_{\rm d} \left(\theta \right)$ is the directional kernel with order $t$ and reference direction $\theta_{\ell}$, reflecting the center of the activation. The hyperparameters include the number of directional kernels $L$ and  the order $t$, reflecting the directional resolution (a larger $t$ indicates more focus in one direction); see Figure~\ref{fig:maze_dir_kernel}.   The kernel coefficient $w_{\ell}$ and the reference direction $\theta_{\ell}$  are the training parameters, which is $\w_{{\Pj}}$ in~\eqref{eq:graph_convolution}.

\textbf{Spatial Kernel.} 
We next consider both direction and distance. The $(i,j)$th element in the graph convolution operator is then,
\vspace{-5mm}
\begin{eqnarray}
\vspace{-5mm}
\label{eq:spatial_kernel}
&& \Pj_{i, j}  \ = \  \Adj_{i, j} \cdot  \sum_{\ell=1}^L w_{\ell}  K^{(d_{\ell}, t,\theta_{\ell} )}_{\rm s} \left( d_{ij},  \theta_{ij} \right), 
\\ \nonumber
&& {\rm where}~ 
K^{(d_{\ell}, t,\theta_{\ell} )}_{\rm s} \left( d, \theta \right) \ = \  
\Id_{ | d - d_{\ell} | \leq \epsilon }
 \left(\frac{1+\cos(\theta - \theta_{\ell})}{2} \right)^t,
\end{eqnarray}
and $d_{ij}$ is the distance between the $i$th and the $j$th nodes, which can be computed through the node embeddings $\X_i, \X_j  \in \R^2$, $K^{(d_{\ell}, t, \theta_{\ell} )}_{\rm s} \left( d, \theta \right)$ is the spatial kernel with reference distance $d_{\ell}$ and reference direction $\theta_{\ell}$ and the indicator function $\Id_{ | d - d_{\ell} | \leq \epsilon } = 1$ when $| d - d_{\ell} | \leq \epsilon$ and $0$, otherwise. The hyperparameters include the number of directional kernels $L$,  the order $t$,  the reference distance $d_{\ell}$ and the distance threshold $\epsilon$. The kernel coefficient $w_{\ell}$ and the reference direction $\theta_{\ell}$ are training parameters ($\w_{{\Pj}}$ in~\eqref{eq:graph_convolution}).

\textbf{Embedding-based Kernel.}  In the directional kernel and spatial kernel, we manually design the kernel and provide hints for GVIN to learn useful direction-distance patterns. Now we directly feed the node embeddings and allow GVIN to automatically learn
implicit hidden factors for general planning. The $(i,j)$th element in the graph convolution operator is then, 
\begin{equation}
\label{eq:emb_kernel}
\Pj_{i, j}  \ = \  \frac{ (\Id_{i=j} + \Adj_{i,j}) }{  \sqrt{ \sum_{k} (1+ \Adj_{k,j}) \sum_k (1+\Adj_{i, k} ) } } \cdot K_{\rm emb} \left( \X_i, \X_j \right),
\end{equation}
where the indicator function $\Id_{ i=j} = 1$ when $i=j$, and $0$, otherwise,  and the embedding-based kernel function is $K_{\rm emb} \left( \X_i, \X_j \right) = {\rm mnnet} \left( \,[   \X_i - \X_j  \,] \right)$, with mnnet$(\cdot)$ is a standard multi-layer neural network. The training parameters $\w_{{\Pj}}$ in~\eqref{eq:graph_convolution} are the weights in the multi-layer neural network.  In practice, when the graph is weighted,  we may also include the graph adjacency matrix $\Adj_{i,j}$ as the input of the multi-layer neural network.

\begin{myThm}
The proposed three kernel functions, the directional kernel, the spatial kernel and the embedding-based kernel,  are shift invariant.
\end{myThm}
The proof follows from the fact that those kernels use only the direction, distance and the difference between two node embeddings, which only depend on the relative position. 


\begin{table*}[h]
	\footnotesize
	\begin{center}
		\begin{tabular}{@{}l>{}c>{}c>{}c>{}c>{}c>{}c}
			\toprule
			& \multicolumn{2}{>{}c}{\bf VIN}    
			& \multicolumn{4}{>{}c}{\bf GVIN}      \\
			& \multicolumn{1}{>{}c}{\bf Action-value}    
			& \multicolumn{1}{>{}c}{\bf State-value}     
			& \multicolumn{2}{>{}c}{\bf Action-value}    
			& \multicolumn{2}{>{}c}{\bf State-value}   \\
			&  &   & dir-aware  & unaware  &  dir-aware  & unaware    \\        
			\midrule \addlinespace[1mm]
			{ Prediction accuracy} &   $95.00\%$  &  $95.00\%$    &   ${ \textbf{95.20\%}}$  &   $92.90\%$   &  $94.40\%$  &  $94.80\%$    \\
			{ Success rate}            &   $99.30\%$  &  $99.78\%$  &  ${ \textbf{99.91\%}}$  &   $98.60\%$   &  $99.57\%$  &  $99.68\%$    \\
			{ Path difference}         &   $0.089$  &  $0.010$  &  $\textbf{0.004}$  &  $0.019$   &  $0.013$  &  $0.015$    \\
			{ Expected reward}       &  $0.963$  & $0.962$  & ${\textbf{0.965}}$  &  $0.939$   &  $0.958$  &  $0.960$    \\
			\bottomrule
		\end{tabular}
	\end{center}
	\caption{\label{table:maze} 2D Maze performance comparison for VIN and GVIN. GVIN achieves similar performance with VIN for 2D mazes ($16 \times 16$); state-value imitation learning achieves similar performance with action-value imitation learning.  }
\end{table*}

\subsection{Training via Reinforcement Learning}
\label{sec:train_rl}

We train GVIN through episodic $Q$-learning, a modified version of $n$-step $Q$-learning. The difference between episodic $Q$-learning and the $n$-step $Q$-learning is that the $n$-step $Q$-learning has a fixed episode duration and
 updates the training weights after $n$ steps; while in episodic $Q$-learning, each episodic terminates when the agent reaches the goal or the maximum step threshold is reached, and we update the trainable weights after the entire episode. During experiments, we found that for both regular and irregular graphs, the policy planned by the original $Q$-learning keeps changing and does not converge due to the frequent updates. Similar to the Monte Carlo algorithms~\cite{sutton1998reinforcement}, episodic $Q$-learning first selects actions by using its exploration policy until the goal is reached. Afterwards, we accumulate the gradients during the entire episode and then update the trainable weights, allowing the agent to use a stable plan to complete an entire episode. This simple change greatly improves the performance (see Section ~\textit{\nameref{sec:maze}}). The pseudocode for the algorithm is presented in Algorithm~\ref{alg:q_learning} (Supplementary~\textit{\nameref{sec:alg_table}}).


\section{Experimental Results}
\label{sec:exp}
  \vspace{-1mm}
 In this section, we evaluate the proposed method on three types of graphs: 2D mazes, synthesized irregular graphs and real road networks. We first validate that the proposed GVIN is comparable to the original VIN for 2D mazes, which have regular lattice structure. We next show that the proposed GVIN automatically learns the concepts of direction and distance in synthesized irregular graphs through the reinforcement learning setting (without using any ground-truth labels). Finally, we use the pre-trained GVIN model to plan paths for the Minnesota road network and Manhattan street network. Additional experiment parameter settings are listed in the Supplementary~\textit{\nameref{sec:exp_setting}}.

\subsection{Revisting 2D Mazes}
\label{sec:maze}
\begin{figure*}[h]
  \begin{center}
      \begin{tabular}{cccc}
    \includegraphics[width=0.24\textwidth]{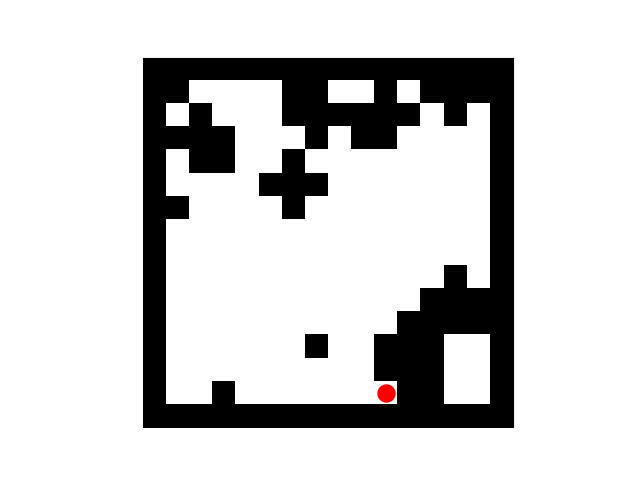}
    & 
    \includegraphics[width=0.24\textwidth]{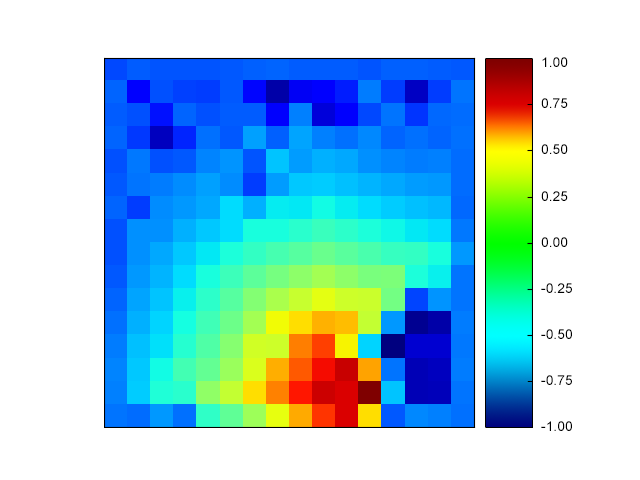}
    &
  \includegraphics[width=0.23\textwidth]{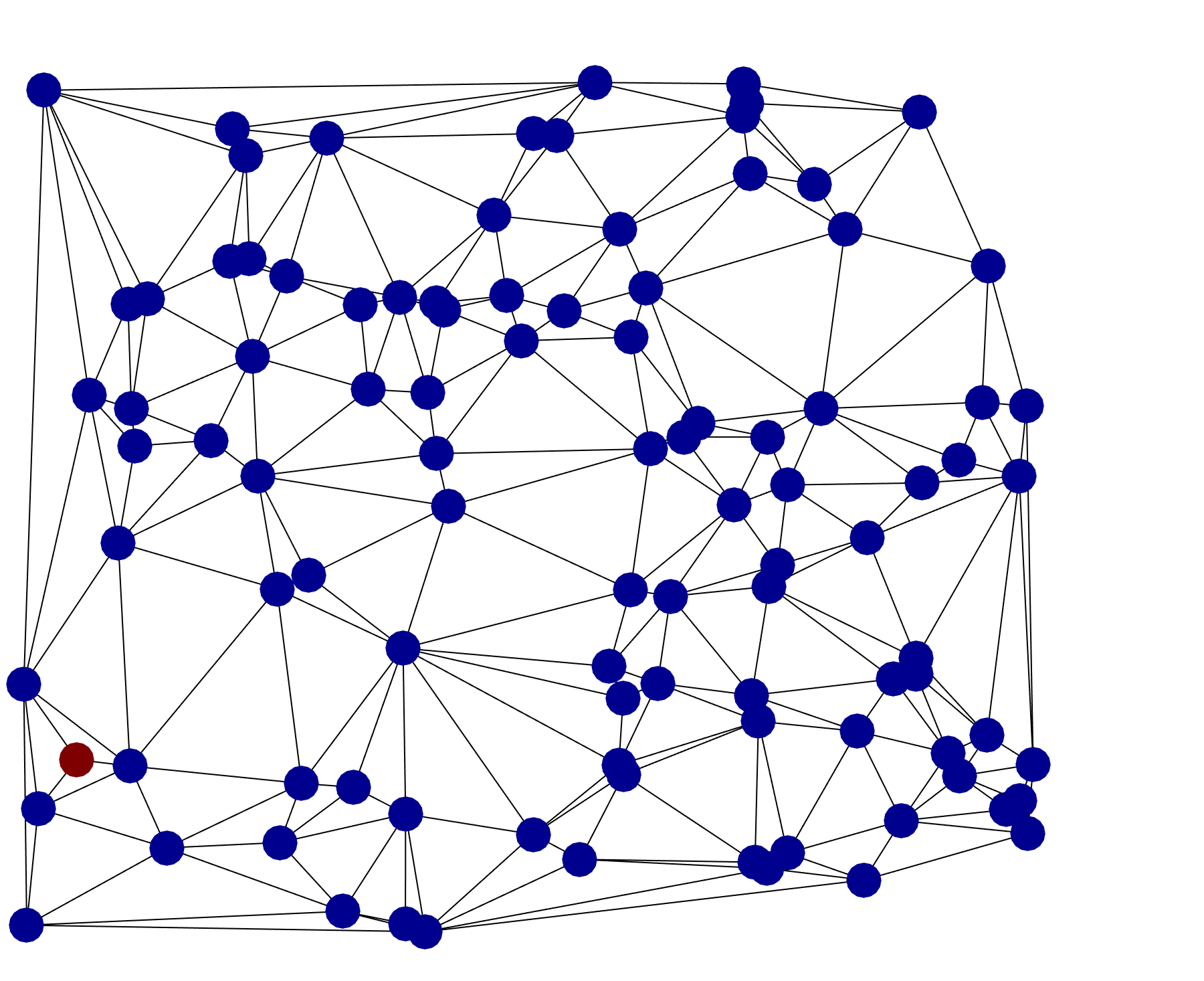}
    & 
    \includegraphics[width=0.23\textwidth]{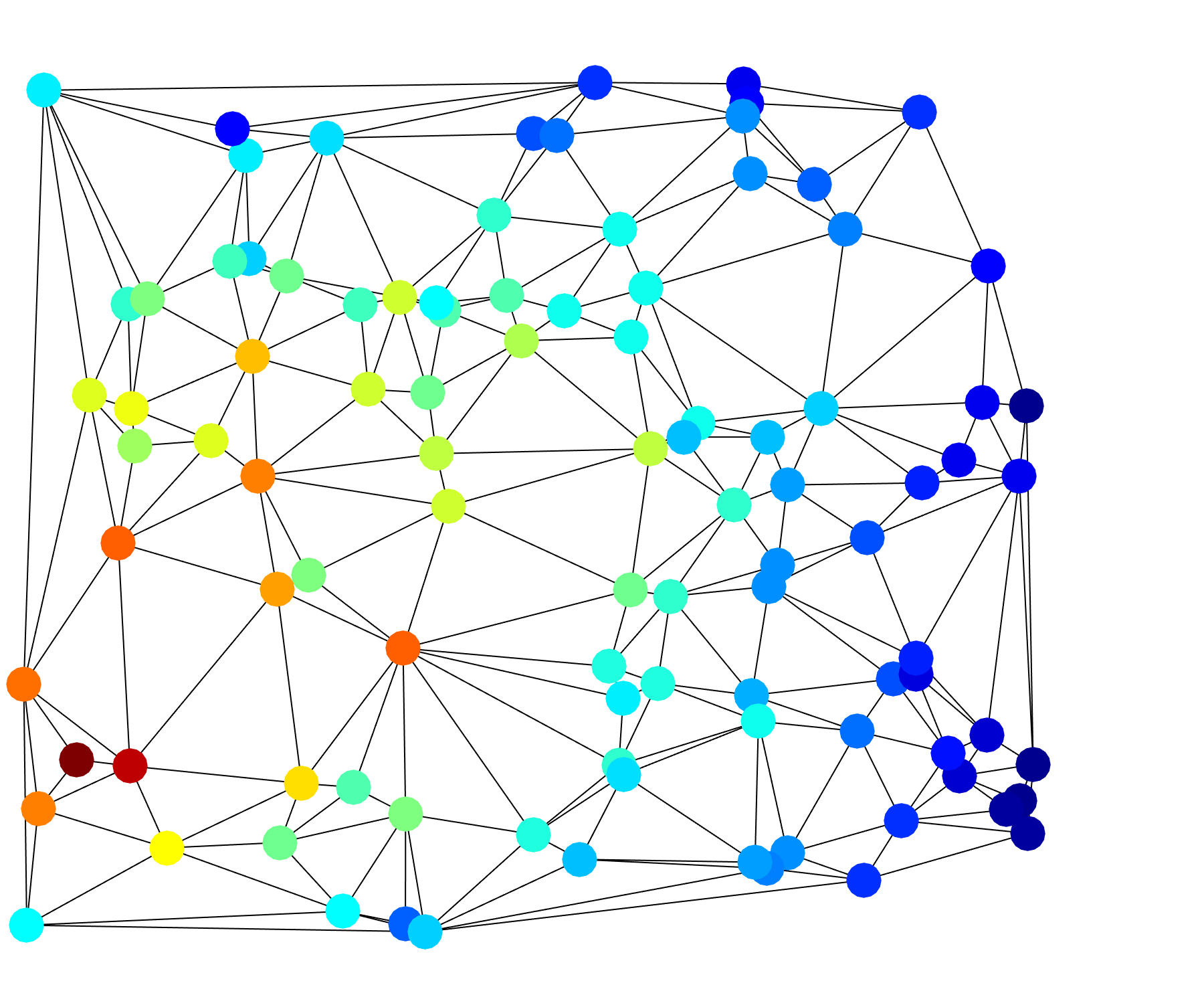}
    \\
    {\small (a) Input map}  &  {\small (b) Value map} &
     {\small (c) Input map}  &  {\small (d) Value map}
    \\
    {\small (2D mazes). }  &  {\small (2D mazes). } &
     {\small (irregular graph). }  &  {\small (irregular graph). }
    \end{tabular}
  \end{center}
  \caption{\label{fig:maze_ir}  value map visualization on regular and irregular graph.}

\end{figure*}

Given a starting point and a goal location, we consider planning the shortest paths for 2D mazes; see Figure~\ref{fig:maze_ir}(a) (Supplementary) as an example.  We generate $22,467$  2D mazes ($16 \times 16$) using the same scripts\footnote{https://github.com/avivt/VIN} that VIN used. We use the same configuration as VIN ($6/7$ data for training and $1/7$ data for testing). Here we consider four comparisons: VIN vs. GVIN, action-value based imitating learning vs. state-value based imitating learning, direction-guided GVIN vs. unguided GVIN, and reinforcement learning.

Four metrics are used to quantify the planning performance, including~\emph{prediction accuracy}---the probability of taking the ground-truth action at each state (higher means better);~\emph{success rate}---the probability of successfully arriving at the goal from the start state without hitting any obstacles (higher means better);~\emph{path difference}---the average length difference between the predicted path and the ground-truth path (lower means better); and~\emph{expected reward}---the average accumulated reward (higher means better). The overall testing results are summarized in Table~\ref{table:maze}.

\mypar{VIN vs. GVIN}
GVIN performs competitively with VIN (Table~\ref{table:maze}), especially when GVIN uses direction-aware action-value based imitation learning (4th column in Table~\ref{table:maze}), which outperforms the others for all four metrics. Figure \ref{fig:maze_ir}(b) (Supplementary) shows the value map learned from GVIN with direction-unaware state-value based imitation learning.
We see negative values (in blue) at obstacles and positive values (in red) around the goal, which is similar to the value map that VIN reported in \cite{tamar2016value}. 

\begin{figure*}[h]
  \begin{center}
      \begin{tabular}{cccc}
    \includegraphics[width=0.21\textwidth]{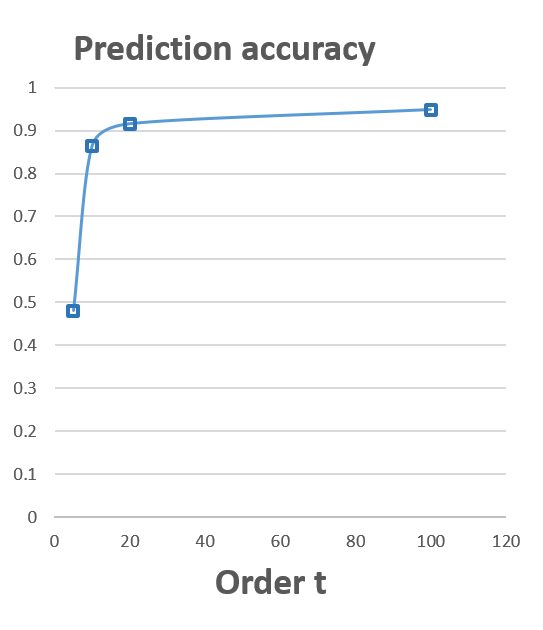}
    & 
    \includegraphics[width=0.21\textwidth]{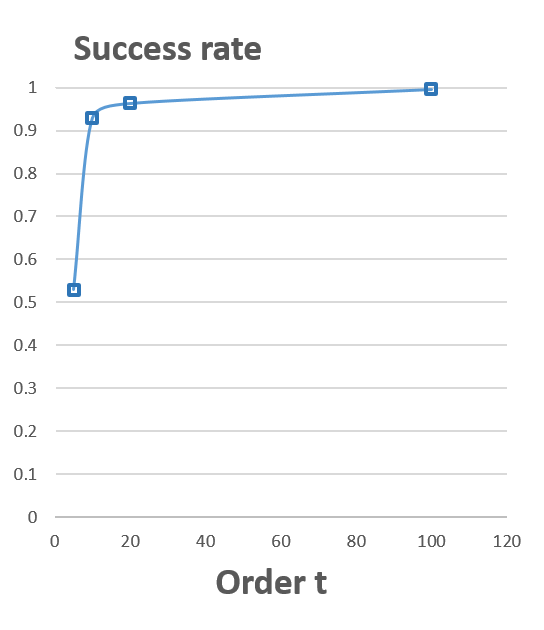}
    &
   \includegraphics[width=0.21\textwidth]{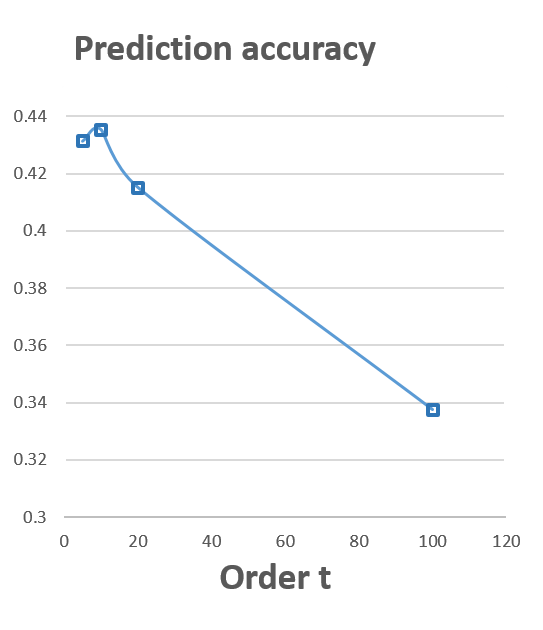}
    & 
    \includegraphics[width=0.21\textwidth]{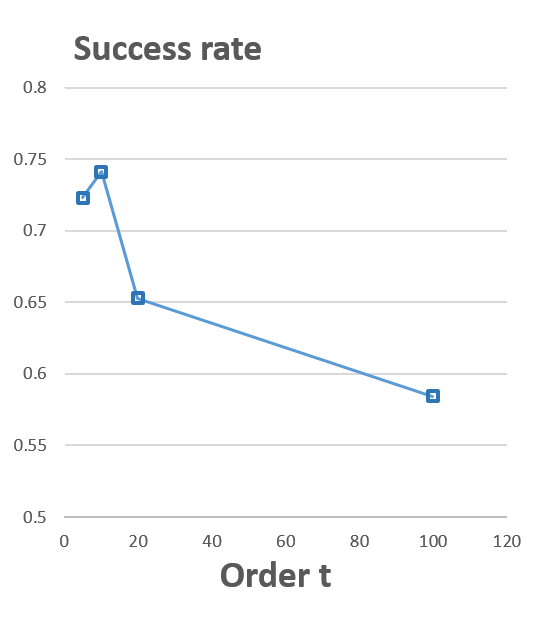}
    \\
    {\small (a)  Prediction accuracy }  &  {\small (b)  Success rate} &
     {\small (c)  Prediction accuracy }  &  {\small (d) Success rate }
     \\
     {\small in 2D mazes.  }  &   {\small in 2D mazes.  }  &  {\small in irregular graphs  }  &  {\small in irregular graphs. }
    \end{tabular}
  \end{center}
  \caption{\label{fig:dir}  Kernel direction order influences the planning performance in both regular and irregular graphs.  }

\end{figure*}

\mypar{Action-value vs. State-value}
VIN with action-value imitation learning slightly outperforms VIN with state-value imitation learning. Similarly, GVIN with action-value based imitation learning slightly outperforms GVIN with state-value based imitation learning. The results suggest that our action approximation method (Section~\textit{\nameref{sec:framework}}) does not impact the performance while maintaining the ability to be extended to irregular graphs.

\begin{figure}[h]
	\begin{center}
		\begin{tabular}{cc}
			\includegraphics[width=0.21\textwidth]{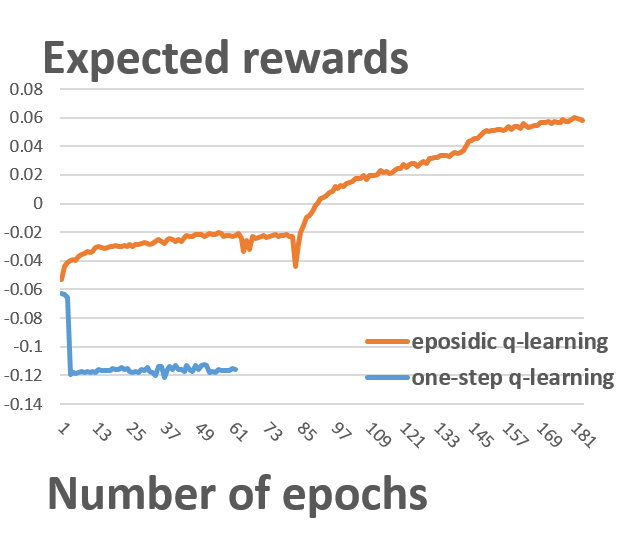}
			& 
			\includegraphics[width=0.21\textwidth]{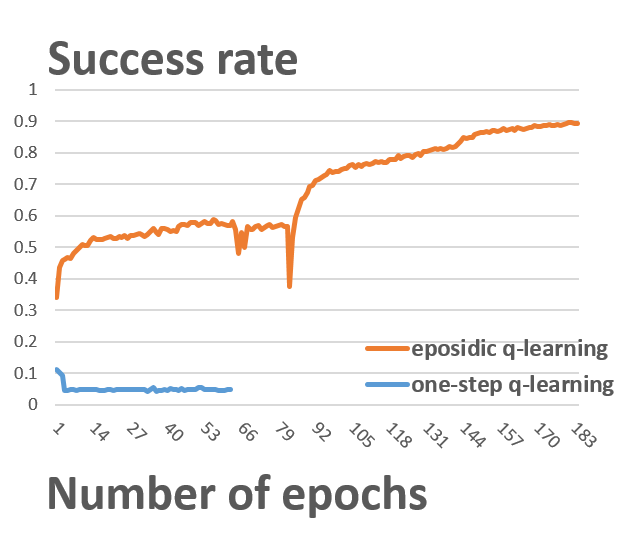}
			\\
			{\small (a) Expected rewards. } & {\small (b) Success rate.}
		\end{tabular}
	\end{center}
	\caption{\label{fig:maze_rl} $Q$- vs. episodic $Q$-learning on $16\times16$~Maze. }
	\vspace{-1mm}
\end{figure} 


\begin{table*}[ht]
\footnotesize
 \begin{center}
 \makebox[\textwidth]{%
 \begin{tabular}{@{}l>{}c>{}c>{}c>{}c>{}c>{}c>{}c>{}c>{}c}
      \toprule
& \multicolumn{1}{>{}c}{\bf VIN}
& \multicolumn{1}{>{}c}{\bf MACN}
& \multicolumn{2}{>{}c}{\bf Directional Kernel}    
& \multicolumn{2}{>{}c}{\bf Spatial Kernel}      
& \multicolumn{3}{>{}c}{\bf Embedding-based Kernel}      \\
      &   & ($36$ nodes) & dir-aware  & unaware   & dir-aware  & unaware  &   train $100$ (IL)  & train $10$ (IL) & train $10$ (RL) \\        
      \midrule \addlinespace[1mm]
{ Prediction acc.}  &  $26.57\%$  & $78\%$ &   $41.50\%$  &  $41.51\%$    &   $ 57.45\%$  &   $57.90\%$   &  $\textbf{58.90\%}$  & $56.14\%$  & $50.90\%$ \\
{ Success rate}     &  $10.29\%$  & $89.4\%$ &   $34.75\%$  &  $65.30\%$  &  $96.56\%$  &   $97.17\%$   &  $97.34\%$  & $6.73\%$ & $\textbf{100\%}$  \\
{ Path diff.}       &   $0.992$  & - &  $0.175$  &  $0.141$  &  $0.082$  &  $0.082$   &  $\textbf{0.079}$    & $0.041$ & $0.14$8\\
{ Expected reward}  &   $-0.905$  & - & $0.266$  & $0.599$  & $  0.911$  &  $0.917$ & $0.922$ & $-0.03$ & $\textbf{0.943}$\\
      \bottomrule
    \end{tabular}}
  \end{center}
  \caption{\label{table:irregular} The performance comparison amongst VIN and three different kernels of GVIN. All experiments except MACN~\cite{khan2017memory} are tested on $100$-node irregular graphs. Note the last column is trained using episodic $Q$-learning. IL and RL stands for imitate learning and reinforcement learning, respectively. Under similar experimental settings, MACN achieves an $89.4\%$ success rate for $36$-node graphs, while GVIN achieves a $97.34\%$ success rate for $100$-node graphs. The details about training VIN on irregular graphs  sees Section~\textit{\nameref{sec:irregular_graph}} in the supplementary material. }
\end{table*}

\mypar{Direction-aware GVIN vs. Unaware GVIN}
Direction-aware GVIN slightly outperforms direction-unaware GVIN, which is reasonable because the fixed eight directions are ground truth for regular 2D mazes. It remains encouraging that the GVIN is able to find the ground-truth directions through imitation learning. As shown later, direction-unaware GVIN outperforms direction-aware GVIN in irregular graphs. Figures~\ref{fig:dir}(a) and (b) show that the planning performance improves as the kernel exponential $t$ in~\eqref{eq:dir_kernel} increases due to the resolution in the reference direction being low when $t$ is small. Figure~\ref{fig:maze_dir_kernel} (in Supplementary) compares the kernel with the same reference direction, but two different kernel orders. When $t =5$, the kernel activates wide-range directions; when $t = 100$, the kernel focuses on a small-range directions and has a higher resolution.

\mypar{Reinforcement Learning}
We also examine the performance of episodic $Q$-learning (Section~\textit{\nameref{sec:train_rl}}) in VIN. Table~\ref{table:rl_maze} (Supplementary) shows that the episodic $Q$-learning algorithm outperforms the training method used in VIN (TRPO + curriculum learning). For the results reported in Table~\ref{table:rl_maze}, we were able to train the VIN using our algorithm (episodic Q-learning) in just 200 epochs, while TRPO and curriculum learning took 1000 epochs to train VIN, as reported in~\cite{tamar2016value} (both algorithms used the same settings). 
As shown in Figure~\ref{fig:maze_rl}, the episodic $Q$-learning algorithm shows faster convergence and better overall performance when compared with $Q$-learning.

\subsection{Exploring Irregular Graphs}
\label{exp:ir}

\begin{table*}[h]
  \footnotesize
  \begin{center}
     \begin{tabular}{@{}l>{}c>{}c>{}c|>{}c>{}c>{}c>{}c}
      \toprule
& \multicolumn{3}{>{}c}{\bf Minnesota}  
& \multicolumn{3}{>{}c}{\bf New York City}  \\
& \multicolumn{1}{>{}c}{\bf Optimal}       
& \multicolumn{1}{>{}c}{\bf $|\V|=100$}    
& \multicolumn{1}{>{}c}{\bf $|\V|=10$}      
& \multicolumn{1}{>{}c}{\bf Optimal}       
& \multicolumn{1}{>{}c}{\bf $|\V|=100$}    
& \multicolumn{1}{>{}c}{\bf $|\V|=10$}      
\\  
      \midrule \addlinespace[1mm]
{ Prediction Accuracy}      &   $100\%$ &   $78.37\%$  &  $78.15\%$  &   $100\%$ & $78.66\%$ & $79.11\%$\\      
{ Success rate}           &   $100\%$ &   $100\%$  &  $100\%$  &   $100\%$ & $100\%$ & $100\%$ \\
{ Path difference}      &   $0.0000$   &   $0.1069$  &  $0.1025$  &   $0.0000$ &  $0.03540$  & $0.0353$\\
{ Expected reward}     &   $0.96043$  &  $0.95063$  & $0.95069$ & $0.97279$ & $0.97110$ & $0.97136$\\
      \bottomrule
    \end{tabular}
  \end{center}
  \caption{\label{table:minnesota} Performance comparison on Minnesota and New York City street map data using GVIN. $|\V|=100$ is trained on $100$-node graphs and $|\V|=10$ is trained on $10$-node graphs.}
\end{table*}

We consider four comparisons in the following experiments: Directional kernel vs. Spatial kernel vs. Embedding-based kernel, direction-aware vs. direction-unaware, scale generalization, and reinforcement learning vs. imitation learning. We use the same performance metrics as the previously discussed 2D maze experiments.

\mypar{Directional Kernel vs. Spatial Kernel vs. Embedding-based Kernel} We first train the GVIN via imitation learning. Table~\ref{table:irregular} shows that the embedding-based kernel outperforms the other kernel methods in terms of both action prediction and path difference (5th column in Table~\ref{table:irregular}), indicating that the embedding-based kernel captures the edge weight information (distance) within the neural network weights better than the other methods. The spatial kernel demonstrates higher accuracy and success rate when compared with the directional kernel, which suggests the effectiveness of using bin sampling. The \textit{direction-unaware} method shows slightly better results for the spatial kernel, but has a larger success rate gain for the directional kernel. Figure~\ref{fig:maze_ir}(d) (Supplementary) shows the visualization of the learned value map which shares similar properties with the regular graph value map. We also train VIN (1st column) by converting graph to 2D image. As shown in the Table, VIN fails significantly (See Supplementary~\textit{\nameref{sec:exp_setting}}).

\begin{figure}[h]
  \begin{center}
      \begin{tabular}{cc}
    \includegraphics[width=0.21\textwidth]{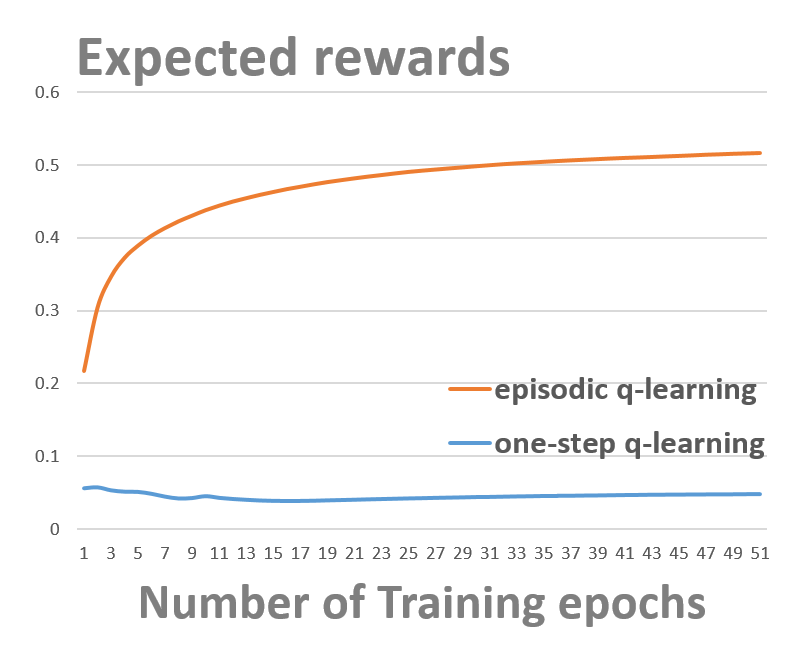}
    & 
    \includegraphics[width=0.21\textwidth]{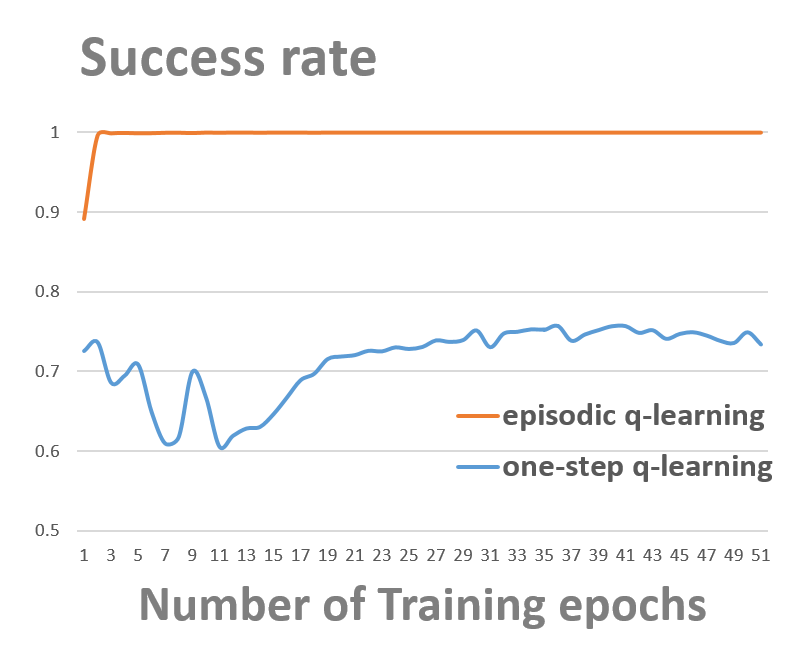}
    \\
    {\small (a) Expected rewards}  &  {\small (b) Success rate}
      \vspace{-3mm}
    \end{tabular}
  \end{center}
  \caption{\label{fig:rl_comp}  $Q$- vs. Episodic $Q$-learning  on irregular graphs.}
  \vspace{-2mm}
\end{figure}

Figures~\ref{fig:dir}(c) and (d) show the planning performance for the irregular domain as the kernel order $t$ in~\ref{eq:dir_kernel} increases. The results show that a larger $t$ in the irregular domain has the opposite effect when compared with the regular domain. The observation is reasonable: in the irregular domain, the direction of each neighbor is extremely variable and a larger kernel order creates a narrower direction range (as seen in Figure\ref{fig:maze_dir_kernel}), thus resulting in information loss.

\mypar{Reinforcement Learning} We then train the GVIN using episodic $Q$-learning to compare with imitation learning. As a baseline, we also train GVIN by using standard deep $Q$-learning techniques, including using an experience replay buffer and a target network. Both networks use the same kernel function (embedding-based kernel) and configurations. Figure~\ref{fig:rl_comp} shows the comparison of the two algorithms' success rate and expected rewards during the training. Clearly, episodic $Q$-learning converges to both a high success rate and a high expected rewards, but the standard deep $Q$-learning techniques fail to achieve reasonable results.

\mypar{Scale Generalization} We also examine the scale generalization by training on $10$-node graphs and then testing on $100$-node graphs using the embedding-based kernel. When GVIN is trained on $10$-node graphs via imitation learning, the performance is significantly hindered as shown in Table~\ref{table:irregular} (6th column). When GVIN is trained using episodic $Q$-learning, Table~\ref{table:irregular} (7th column) shows excellent generalization abilities that outperform all imitation learning based results for success rate and expected rewards. Compared with imitation learning, we also observe the performance decreases for path differences and action prediction.

\mypar{Graph with Edge Weights} We also test how GVIN  handles edge weights. We set the true weighted shortest path to be $\frac{\X_i-\X_j}{W_{ij}}$, where $\X_i-\X_j$ is the distance between two nodes and $W_{ij}$ is the edge weight. As shown in Table~\ref{table:edge_weight}, imitation learning is trained on $100$-node graphs, while reinforcement learning is trained on $10$-node. We also examine the GVIN by excluding edge weights from the input to see if there are any effects on performance. Table~\ref{table:edge_weight} (Supplementary) shows that for reinforcement learning, edge weights slightly help the agent find a more suitable policy; for imitation learning, input edge weights cause a significant failure.

\vspace{-2mm}
\subsection{Validating Real Road Networks}
\vspace{-1mm}
To demonstrate the generalization capabilities of GVIN, we evaluate two real-world maps: the Minnesota highway map, which contains $2642$ nodes representing intersections and $6606$ edges representing roads, and the New York City street map, which contains $5069$ nodes representing intersections and $13368$ edges representing roads. We use the same models trained on the graphs containing $|\V|=100$ and $|\V|=10$ nodes with the embedding-based kernel and using episodic $Q$-learning in Section~\textit{\nameref{exp:ir}}, separately. We normalize the data coordinates between $0$ and $1$, and we set recurrence parameter to $K=200$. We randomly pick start points and goal points 1000 different times. We use the A* algorithm as a baseline. Table~\ref{table:minnesota} shows that both $|\V|=100$ and $|\V|=10$ generalize well on large scale data. The policy could reach the goal position with $100\%$ in the experiments. One sample planned path is shown in Supplementary (Figures~\ref{fig:min_visual} and~\ref{fig:nyc_visual}).


\vspace{-2mm}
\section{Conclusions}
We have introduced GVIN, a differentiable, novel planning module capable of both regular and irregular graph navigation and impressive scale generalization. We also introduced episodic $Q$-learning that is designed to stabilize the training process of VIN and GVIN. The proposed graph convolution may be applied to many other graph-based applications, such as navigation, 3D point cloud processing and molecular analysis, which is left for future works.



\bibliography{./reference}
\bibliographystyle{aaai}


\clearpage

\section{Appendix}

\subsection{Computational Complexity}
\label{sec:complexity}
Let  the input graph $G$ have $|\V|$ nodes and $|\E|$ edges. In the testing phase, since the input graph are commonly sparse, the computational complexities of~\eqref{eq:reward},~\eqref{eq:bellman} and~\eqref{eq:value_map} are $O(|\V|)$, $O(|\E|)$ and $O(|\V|)$ based on sparse computation, respectively.  Therefore, the total computational complexity is $O( |\V| + K(|\E| +|\V| ))$, where $K$ is number of iterations. For a spatial graph, the number of edges is usually proportional to the number of nodes; thus the computational complexity is $O(K|\V|)$, which is scalable to huge graphs.

\subsection{Episodic $Q$-learning }
\label{sec:alg_table}
We highlight the differences between episodic $Q$-learning  and the original $n$-step $Q$-learning in blue, including the initial expected return, the termination condition and the timing of updating the gradient.
\begin{algorithm}
\caption{Episodic $Q$-learning }
\label{alg:q_learning}
\begin{algorithmic}[1]
\State  input graph $G$ and the goal $s_{\rm g}$
\State  initialize global step counter $T=0$
\State initialize GVIN parameters $\w = [\w_\r, \w_{\Pj^{(a)}}]$
\State initialize parameter gradients $\Delta \w$ 
\Repeat ~(\textcolor[rgb]{0, 0.2, 1}{one episode})
    \State clear gradients $\Delta \w \leftarrow 0$
    \State $t = 0$
    \State randomly pick a start node $s_t$
    \Repeat  ~(one action)
        \State take action $a_t$ according to the $\epsilon$-greedy policy based on $\q^{(a)}_{s_t}$
        \State receive reward $r_t$ and new state $s_{t+1}$
        \State $t \leftarrow t+1$
      \Until   \textcolor[rgb]{0, 0.2, 1}{terminal $s_t = s_{\rm g}$ or $t > t_{\rm max}$}
  \State   \textcolor[rgb]{0, 0.2, 1}{$R=0$}
    \For{ $i =  t : -1 : 0 $}
        \State $R \leftarrow r_i + \gamma R$
        \State accumulate gradients wrt $\w: \Delta \w \leftarrow \Delta \w + \frac{\partial(R -  \q^{(a)}_{s_t} )^2}{\partial \w}$
    \EndFor
    \State  \textcolor[rgb]{0, 0.2, 1}{$\w \leftarrow  \w -  \Delta \w$}
    \State  $T = T+1$
\Until $T > T_{\rm max}$
\end{algorithmic}
\end{algorithm}

\subsection{Graph-based Kernel Functions}
\label{sec:kernel}

\textbf{Directional Kernel.} We first consider direction. When we face several roads at an intersection, it is straightforward to pick the one whose direction points to the goal. 
We aim to use the directional kernel to capture edge direction and parameterize the graph convolution operation.

The $(i,j)$th element in the graph convolution operator models the probability of following the edge from $i$ to $j$ ; that is, 
\begin{equation}
\begin{split}
\label{eq:dir_kernel}
\Pj_{i,j}  \ = \  \Adj_{i,j} \cdot \sum_{\ell=1}^L w_{\ell}  K^{(t,\theta_{\ell} )}_{\rm d} \left(\theta_{ij} \right), \\~{\rm where}~ K^{(t,\theta_{\ell} )}_{\rm d} \left(\theta \right) \ = \ \left(\frac{1+\cos(\theta - \theta_{\ell})}{2} \right)^t.
\end{split}
\end{equation}
where  $w_{\ell}$ is kernel coefficient,  $\theta_{ij}$ is the direction of the edge connecting the $i$th and the$j$th nodes, which can be computed through the node embeddings $\X_i \in \R^2$ and $\X_j  \in \R^2$, and $ K^{(t,\theta_{\ell} )}_{\rm d} \left(\theta \right)$ is the directional kernel with order $t$ and reference direction $\theta_{\ell}$, reflecting the center of the activation. The hyperparameters include the number of directional kernels $L$ and  the order $t$, reflecting the directional resolution (a larger $t$ indicates more focus in one direction); see Figure~\ref{fig:maze_dir_kernel}.   The kernel coefficient $w_{\ell}$ and the reference direction $\theta_{\ell}$  are the training parameters, which is $\w_{{\Pj}}$ in~\eqref{eq:graph_convolution}.  Note that the graph convolution operator $\Pj \in \R^{N \times N}$ is a sparse matrix and its sparsity pattern is the same with the input adjacency matrix, which ensures that the computation is cheap.

\begin{figure}[h]
  \begin{center}
      \begin{tabular}{cc} 
    \includegraphics[width=0.19\textwidth]{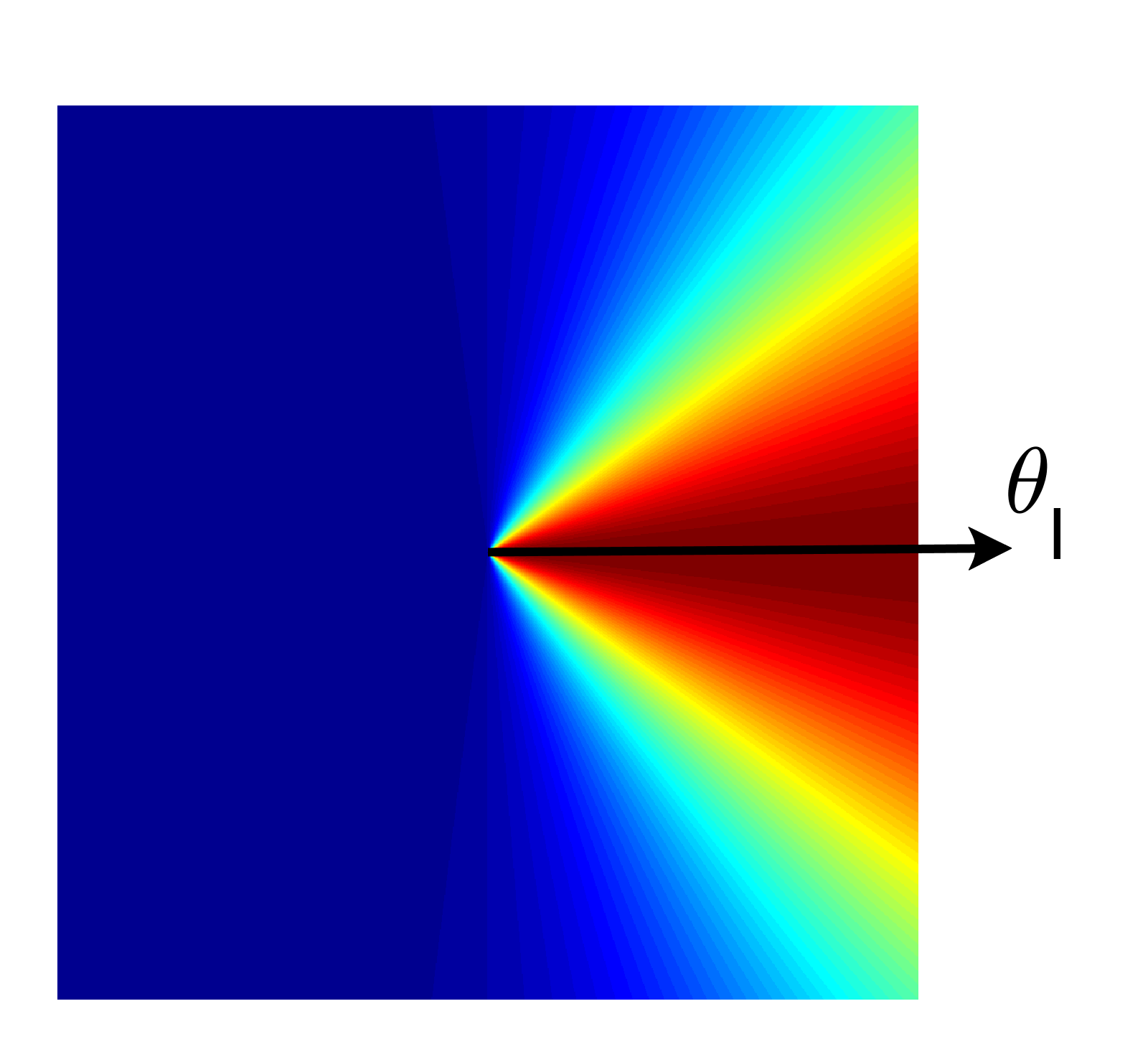}
    & 
    \includegraphics[width=0.19\textwidth]{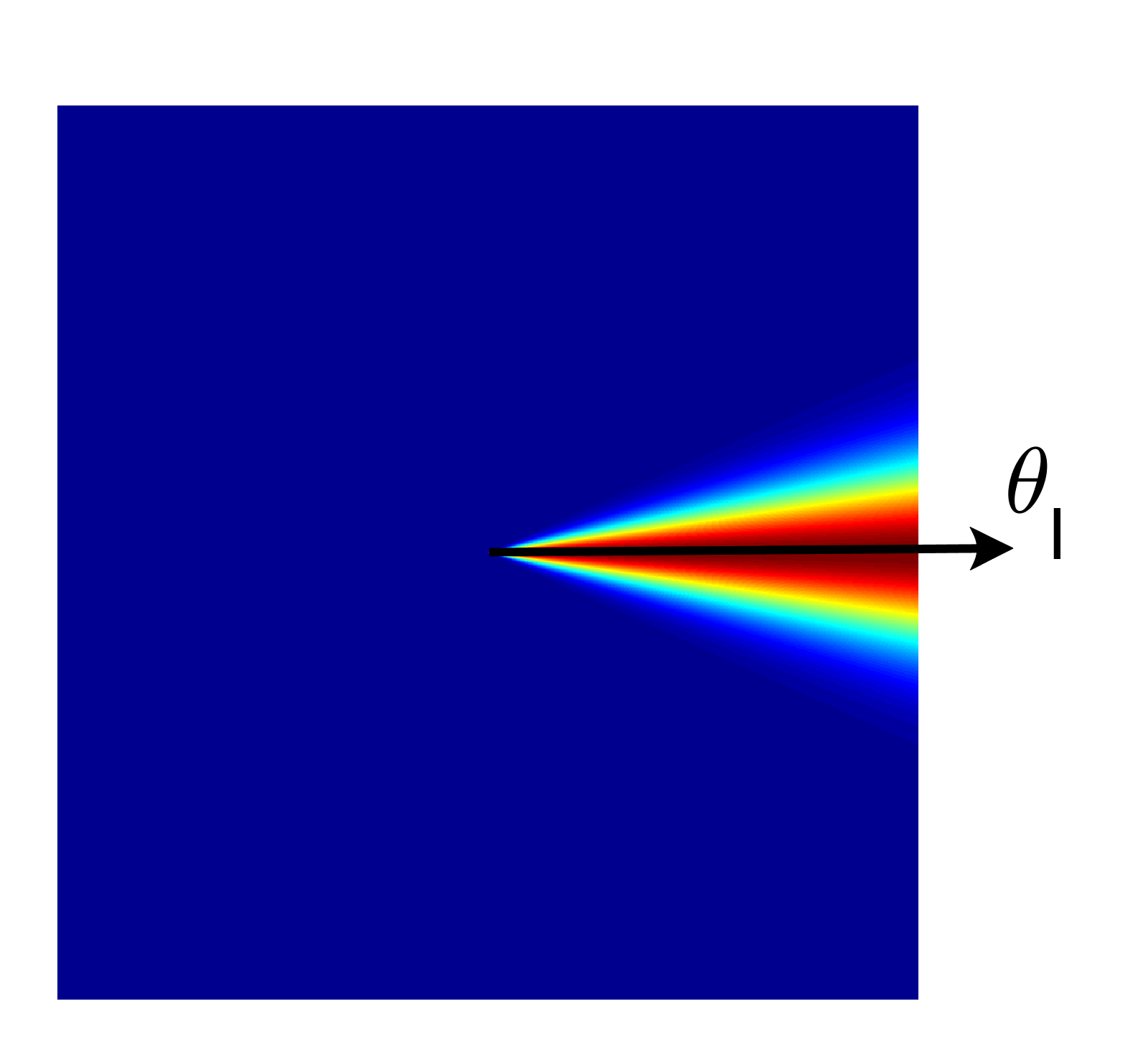}
    \\
    {\small (a) $t = 5$}  &  {\small (b) $t = 100$}
    \end{tabular}
  \end{center}
  \caption{\label{fig:maze_dir_kernel} The directional kernel function activates the areas around the reference direction 
  $\theta_{\ell}$  in the 2D spatial domain. The activated area is more concentrated when $t$ increases. }
\end{figure}

The intuition behind~\eqref{eq:dir_kernel} is that each graph convolution operator represents a unique direction pattern. An edge is a 2D vector sampled from the 2D spatial plane. When the direction of the edge connecting $i$ and $j$ matches one or some of the $L$ reference directions, 
we have a higher probability to follow the edge from $i$ to $j$. In GVIN, we consider several channels. In each channel, we obtain an action value for each node, which represents the matching coefficient of a direction pattern.  The max-pooling operation then selects the most matching direction pattern for each node.

\textbf{Spatial Kernel.} 
We next consider both direction and distance. When all the roads at the current intersection opposite to the goal, it is straightforward to try the shortest one first.  We thus include the edge length into the consideration. The $(i,j)$th element in the graph convolution operator is then, 
\begin{equation}
\begin{split}
\label{eq:spatial_kernel}
\Pj_{i, j}  \ = \   \Adj_{i, j} \cdot  \sum_{\ell=1}^L w_{\ell}  K^{(d_{\ell}, t,\theta_{\ell} )}_{\rm s} \left( d_{ij},  \theta_{ij} \right), 
\\~{\rm where}~ 
K^{(d_{\ell}, t,\theta_{\ell} )}_{\rm s} \left( d, \theta \right) \ = \  
\Id_{ | d - d_{\ell} | \leq \epsilon }
 \left(\frac{1+\cos(\theta - \theta_{\ell})}{2} \right)^t,
\end{split}
\end{equation}
where $d_{ij}$ is the distance between the $i$th and the $j$th nodes, which can be computed through the node embeddings $\X_i \in \R^2$ and $\X_j  \in \R^2$, $K^{(d_{\ell}, t, \theta_{\ell} )}_{\rm s} \left( d, \theta \right)$ is the spatial kernel with reference distance $d_{\ell}$ and reference direction $\theta_{\ell}$ and the indicator function $\Id_{ | d - d_{\ell} | \leq \epsilon } = 1$ when $| d - d_{\ell} | \leq \epsilon$ and $0$, otherwise. The hyperparameters include the number of directional kernels $L$,  the order $t$,  the reference distance $d_{\ell}$ and the distance threshold $\epsilon$. The kernel coefficient $w_{\ell}$ and the reference direction $\theta_{\ell}$ are training parameters, which is $\w_{{\Pj}}$ in~\eqref{eq:graph_convolution}. 

\begin{figure}[h]
  \begin{center}
      \begin{tabular}{cc} 
    \includegraphics[width=0.19\textwidth]{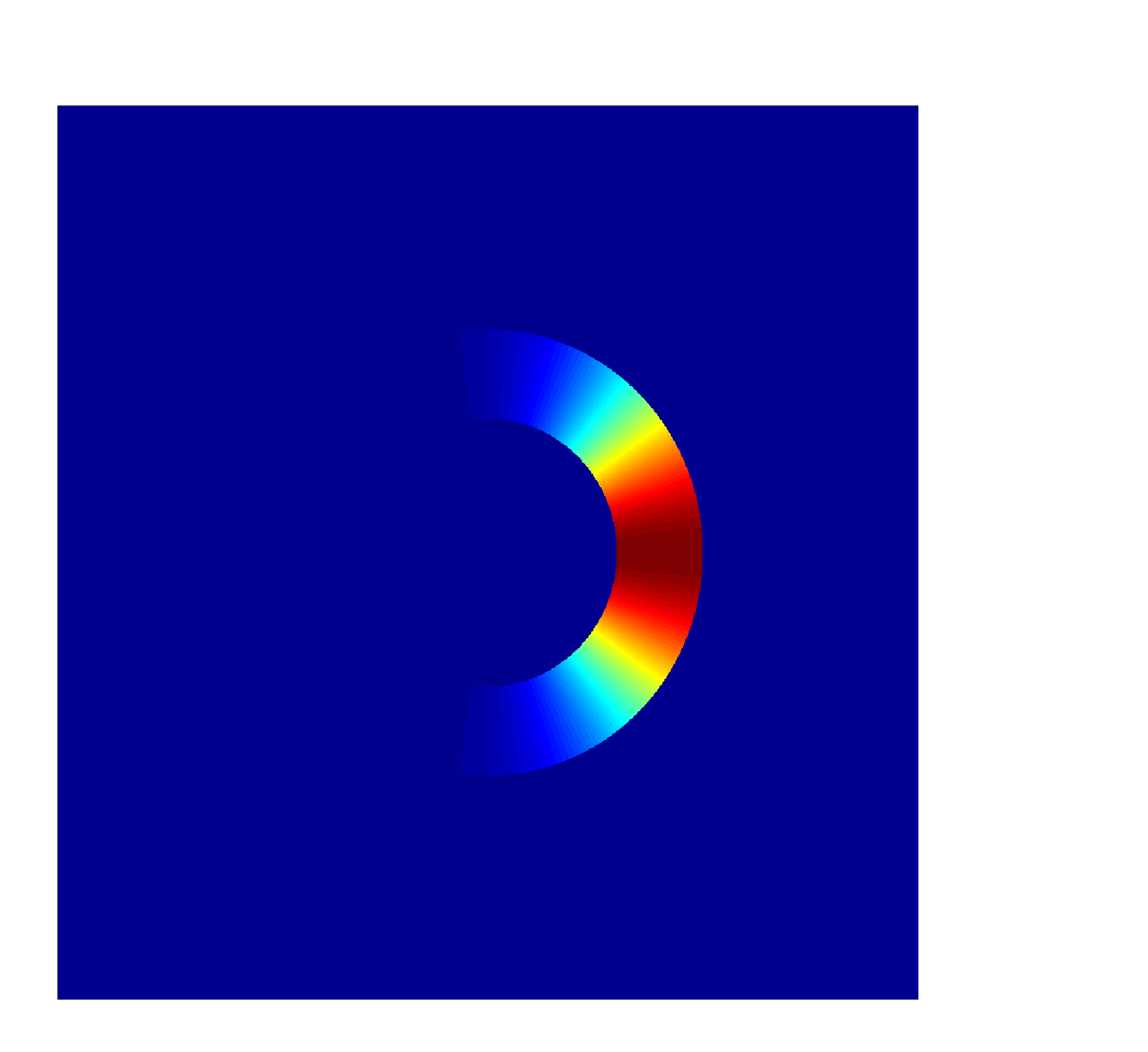}
    & 
    \includegraphics[width=0.19\textwidth]{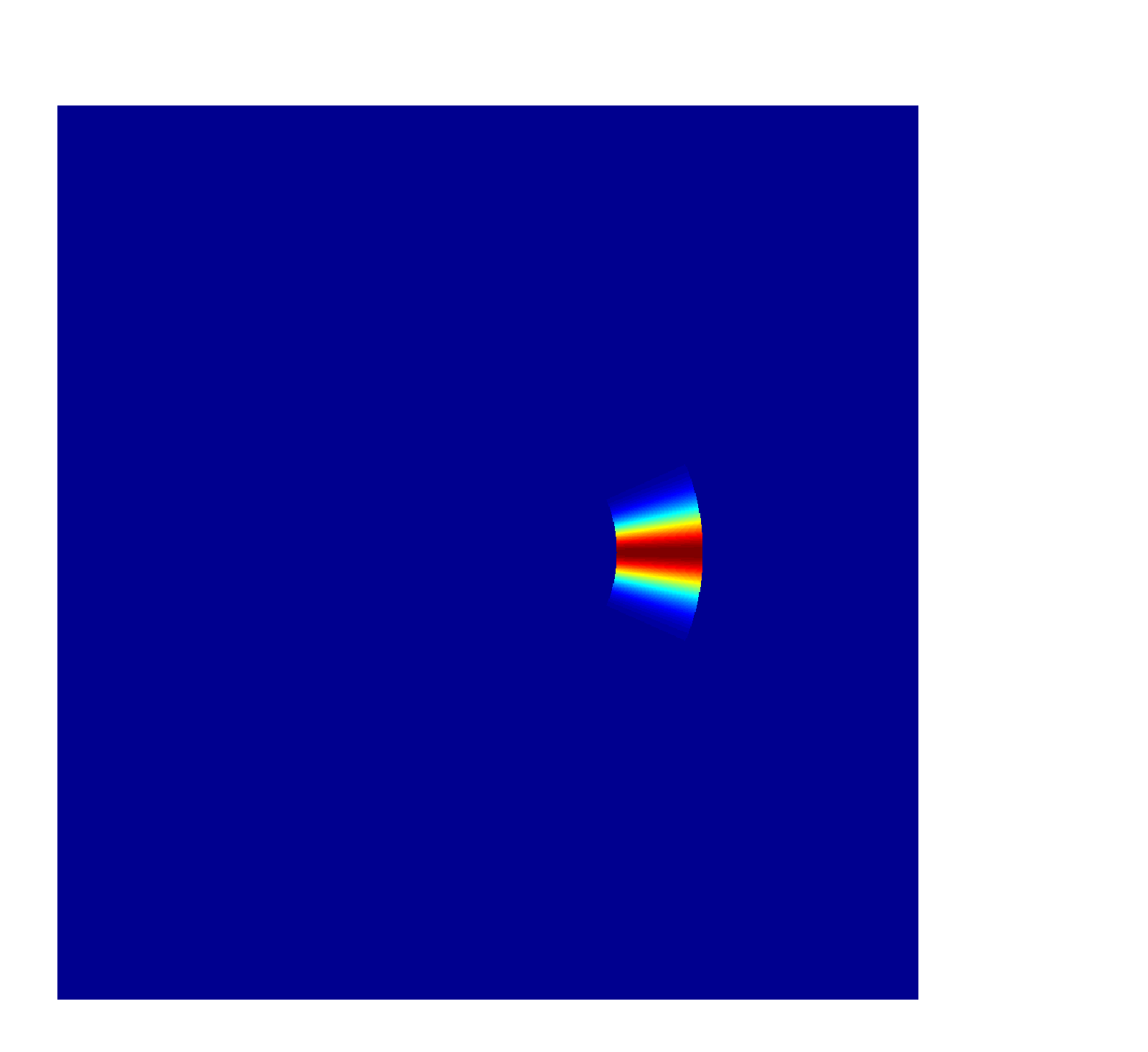}
    \\
    {\small (a) $t = 5$.}  &  {\small (b) $t = 100$.}
    \end{tabular}
  \end{center}
  \caption{\label{fig:maze_dir_kernel} The spatial kernel function activates the areas around the reference direction 
  $\theta_{\ell}$ and reference distance $d_{\ell}$ in the 2D spatial domain.  }
\end{figure}

Compared to the directional kernel, the spatial kernel adds another dimension, distance; in other words, the directional kernel is a special case of the spatial kernel when we ignore the distance.  Each spatial kernel activates a localized area in the direction-distance plane.   With  the spatial kernel, the graph convolution operator~\eqref{eq:spatial_kernel} represents a unique direction-distance pattern; that is, if the direction/distance of the edge connecting $i$ and $j$ matches one or some of the $L$ reference directions and distances, we have a higher probability to follow the edge from $i$ to $j$.

\textbf{Embedding-based Kernel.}  In the directional kernel and spatial kernel, we manually design the kernel and hint GVIN to learn useful direction-distance patterns. Now we directly feed the node embeddings and allow GVIN to automatically learn
implicit hidden factors for general planning. The $(i,j)$th element in the graph convolution operator is then, 
\begin{equation}
\label{eq:emb_kernel}
\Pj_{i, j}  \ = \  \frac{ (\Id_{i=j} + \Adj_{i,j}) }{  \sqrt{ \sum_{k} (1+ \Adj_{k,j}) \sum_k (1+\Adj_{i, k} ) } } \cdot K_{\rm emb} \left( \X_i, \X_j \right),
\end{equation}
where the indicator function $\Id_{ i=j} = 1$ when $i=j$, and $0$, otherwise,  and the embedding-based kernel function is $K_{\rm emb} \left( \X_i, \X_j \right) = {\rm mnnet} \left( \,[ \Adj_{ij},   \X_i - \X_j \,] \right)$, with mnnet$(\cdot)$ is a standard multi-layer neural network. The training parameters $\w_{{\Pj}}$ in~\eqref{eq:graph_convolution} are the weights in the multi-layer neural network.  Note that the graph convolution operator $\Pj \in \R^{N \times N}$ is still a sparse matrix and its sparsity pattern is the same with the input adjacency matrix plus the identity matrix.

In the directional kernel and spatial kernel, we implicitly discretize the space based on the reference direction and distance; that is, for each input pair of given direction and distance,  the kernel function outputs the response based on its closed reference direction and distance. In the embedding-based kernel, we do not set the reference direction and distance to discretize the space; instead, we use the multi-layer neural network to directly regress from an arbitrary edge (with edge weight and embedding representation) to a response value. The embedding-based kernel is thus more flexible than  the directional kernel and spatial kernel and may learn hidden factors.

\subsection{Experiment Settings}
\label{sec:exp_setting}
Our implementation is based on Tensorflow with GPU-enabled platform. All experiments use the standard centered RMSProp algorithm as the optimizer with learning rate $\eta=0.001$~\cite{tieleman2012lecture}. All reinforcement learning experiments use a discount of $\gamma=0.99$, RMSProp decay factor of $\alpha=0.999$, and exploration rate $\epsilon$ annealed linearly from $0.2$ to $0.001$ over the first 200 epochs. 

\subsubsection{2D Mazes}

The experiments are set up as follows. We consider the rules as follows: the agent receives a $+1$ reward when reaching the goal, receives a $-1$ reward when hitting an obstacle, and each movement gets a $-0.01$ reward. To preprocess the input data, we use the same two-layer CNN for both VIN and GVIN, where the first layer involves $150$ kernels with size $3\times3$ and the second layer involves a kernel with size $3\times3$ for output. The transition probability matrix is parameterized by $10$ convolution kernels with size  $3\times3$ in both VIN and GVIN. In GVIN, we use the directional kernel based method as shown in Equations~\ref{eq:dir_kernel} and~\ref{eq:spatial_kernel} and we set $\ell=8$ to represent the eight reference directions. We consider two approaches to initialize the directions $\theta_\ell$. In {direction-aware} approach, we fix $\theta_\ell$ as $[0,\pi/4,\pi/2,...,7\pi/4]$. In the direction-unaware approach, we set $\theta_\ell$ to be weights and train them via backpropagation. We set the recurrence $K$ in GVIN to be $20$ for $16\times16$ 2D mazes. In the regular domain, we set the kernel order $t=100$ to be the default.

\begin{table}[t]
	\footnotesize
	\begin{center}
		\begin{tabular}{lrrrrr}
			\toprule 
			& \multicolumn{1}{c}{\bf TRPO}    
			& \multicolumn{1}{c}{\bf EQL\footnote{represent Episodic Q-learning}}      \\
			\midrule \addlinespace[1mm]
			{ Success rate}            & $82.50\%$ &  $98.67\%$    \\
			{ No. of Epochs}           & $1000$ & $200$ \\
			{ Path difference}         & $N/A$ &  $0.1617$    \\
			{ Expected reward}       & $N/A$ & ${  0.9451}$     \\
			\bottomrule
		\end{tabular}
	\end{center}
	\caption{\label{table:rl_maze} Performance comparison using different training algorithms on the VIN model. The first column is VIN trained by TRPO with curriculum learning reported in~\cite{tamar2016value}, the second column is VIN trained by episodic $Q$-learning.  }
\end{table}

\subsubsection{Irregular Graphs}
\label{sec:irregular_graph}

\begin{table*}[h]
  \footnotesize
  \begin{center}
     \begin{tabular}{@{}l>{}c>{}c|>{}c>{}c}
      \toprule
& \multicolumn{2}{>{}c}{\bf Imitation Learning}  
& \multicolumn{2}{>{}c}{\bf Reinforcement Learning}  \\
& \multicolumn{1}{>{}c}{\bf w edge weight}    
& \multicolumn{1}{>{}c}{\bf w/o edge weight}      
& \multicolumn{1}{>{}c}{\bf w edge weight}      
& \multicolumn{1}{>{}c}{\bf w/o edge weight}      \\        
      \midrule \addlinespace[1mm]
{ Prediction accuracy} &   $62.24\%$  &  $43.12\%$    &   $\textbf{50.38\%}$ &  $50.14\%$ \\
{ Success rate}            &   $3.00\%$  &  $97.60\%$  &  $\textbf{100\%}$ & $100\%$  \\
{ Path difference}         &   $0.020$  &  $0.763$  &  $\textbf{0.146}$ & $0.151$  \\
{ Expected reward}       &  $-0.612$  & $0.771$  & $\textbf{0.943}$  & $0.940$   \\
      \bottomrule
    \end{tabular}
  \end{center}
  \caption{\label{table:edge_weight} Performance comparison for testing weighted graphs. Imitation learning is trained on $100$-node irregular graphs while reinforcement learning is trained on $10$-node irregular graphs. }
\end{table*}

We evaluate our proposed methods in Section~\textit{\nameref{sec:graph_convolution}} for the irregular domain. Our experimental domain is a synthetic data which consists of $N=10000$ irregular graphs, in which each graph contains $100$ nodes. We follow the standard rules of random geometric graphs to generate our irregular graphs. Specifically, we generate $|\V|$ vertices with random coordinates in the box $[0, 1]^2$ and connects a pair of two vertices with an edge when the distance between two vertices is smaller than a certain threshold. For each node in the graph, we define the coordinates which represent its spatial position ranged between $0$ and $1$. The dataset is split into $7672$ graphs for training and $1428$ graphs for testing. Additionally, to exam whether GVIN could handle weighted graphs, we also generated a synthetic dataset consisting of $100$ node graphs and partitioned $42857$ graphs for training and $7143$ graphs for testing.

For the directional kernel and spatial kernel, we set the number of reference directions to $\ell=8$ and kernel order $t=20$ to be default values for Equations ~\ref{eq:dir_kernel} and ~\ref{eq:spatial_kernel}. We also set $\theta_\ell$ to be $0$ to $2\pi$ with an interval of $\pi/4$ for \textit{direction-aware} mode and we set $\theta_\ell$ to be trainable weights for \textit{direction-unaware} mode. For the spatial kernel function, we set the number of bins $d_\ell$ to be $10$ in Equation \ref{eq:spatial_kernel}. In the embedding-based kernel, we use three layers of fully connected neural networks (32-64-1), where each layer uses $ReLU(\cdot)=\max(0,\cdot)$ as its activation function. The neural network are initialized with zero-mean and 0.01 derivation. or all three kernel methods, we set the graph convolution channel number to be $10$ and $K=40$ for recurrence. 

\mypar{Training VIN on Irregular Graphs}
To show the strong generalization of GVIN, we evaluate the VIN on irregular graph by converting graph data format to 2D image. Each testing set contains reward map and obstacle map that sizes $100\times100$ pixels. We use pre-trained weights from $28\times28$ maze with all parameters tuned to be highest performance. To make the training and testing consistent, we set the rewards map and obstacle map the same settings as 2D maze: vertices and edges are marked as free path (value set to be $0$), while the other area are marked as obstacles (value set to be $1$). The edge path is generated via Bresenham's line algorithm~\cite{bresenham1977linear}. We set recurrence $K$ to be $200$ so that the value iteration could cover the whole map.

\clearpage

\begin{figure*}[h]
  \begin{center}
      \begin{tabular}{cc} 
    \includegraphics[width=0.49\textwidth]{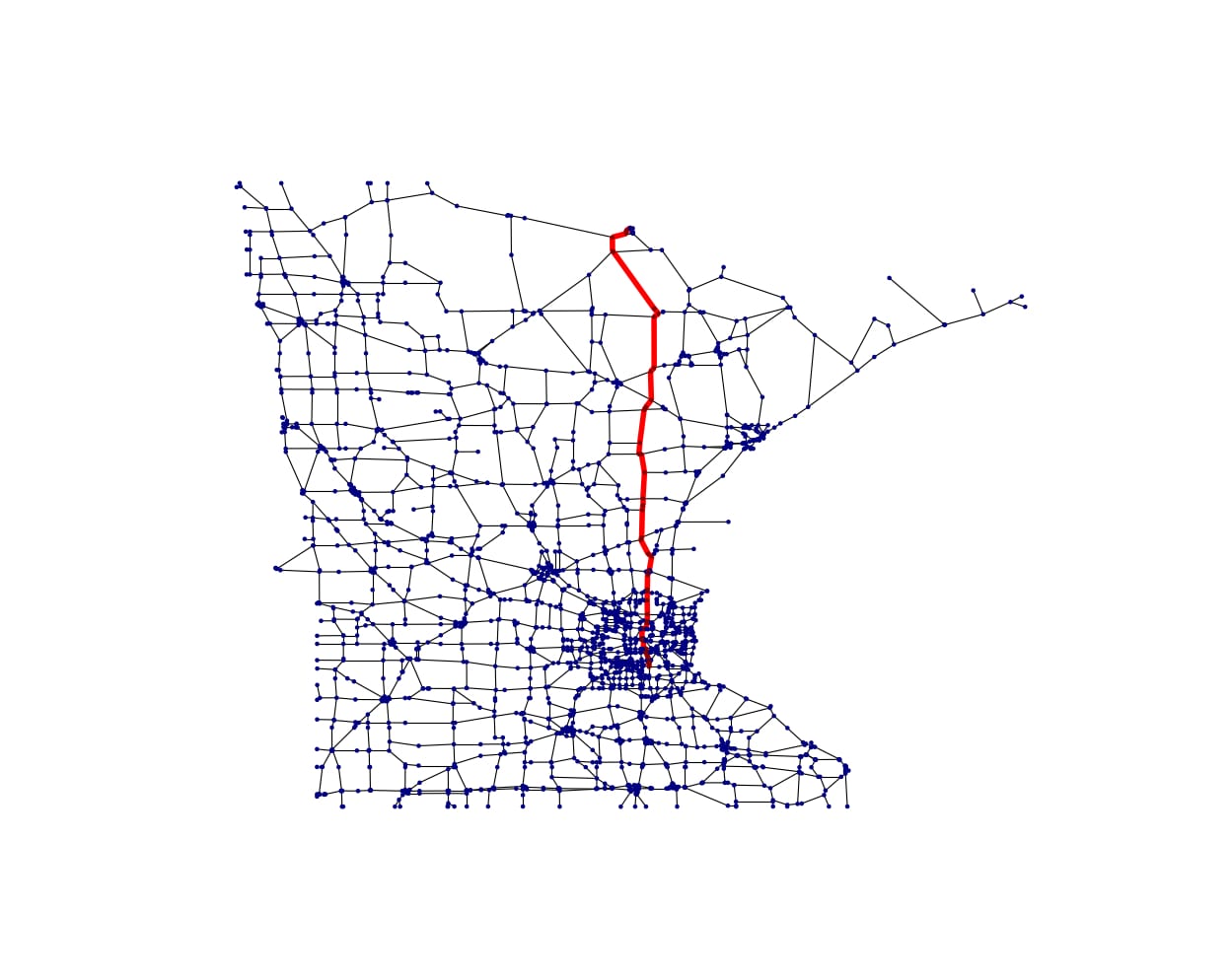}
    & 
    \includegraphics[width=0.49\textwidth]{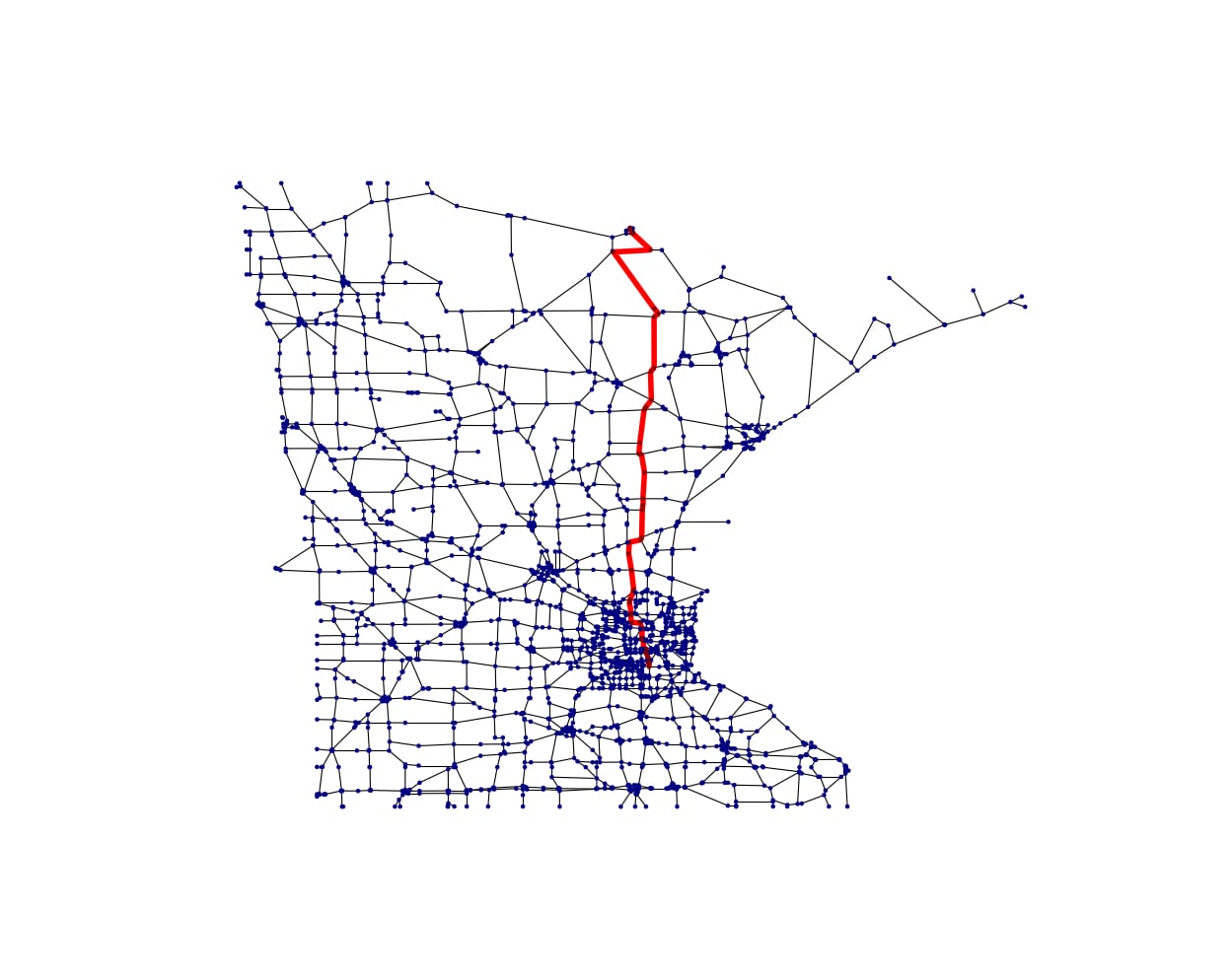}
    \\
    {\small (a) Ground-Truth}  &  {\small (b) GVIN prediction}
    \end{tabular}
  \end{center}
  \caption{\label{fig:min_visual} Sample planning trajectories on Minnesota highway map.}
\end{figure*}

\begin{figure*}[h]
  \begin{center}
      \begin{tabular}{cc} 
    \includegraphics[width=0.49\textwidth]{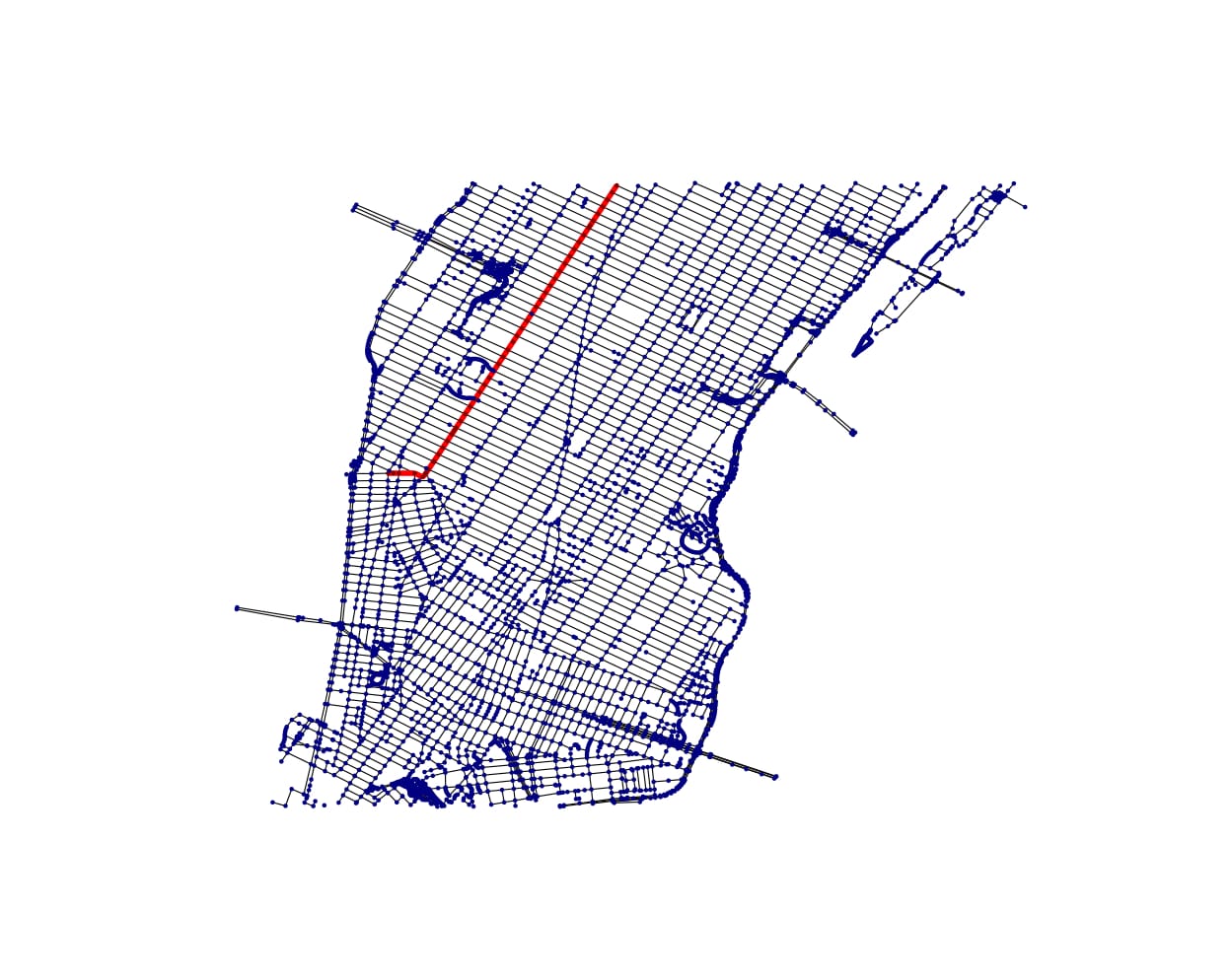}
    & 
    \includegraphics[width=0.49\textwidth]{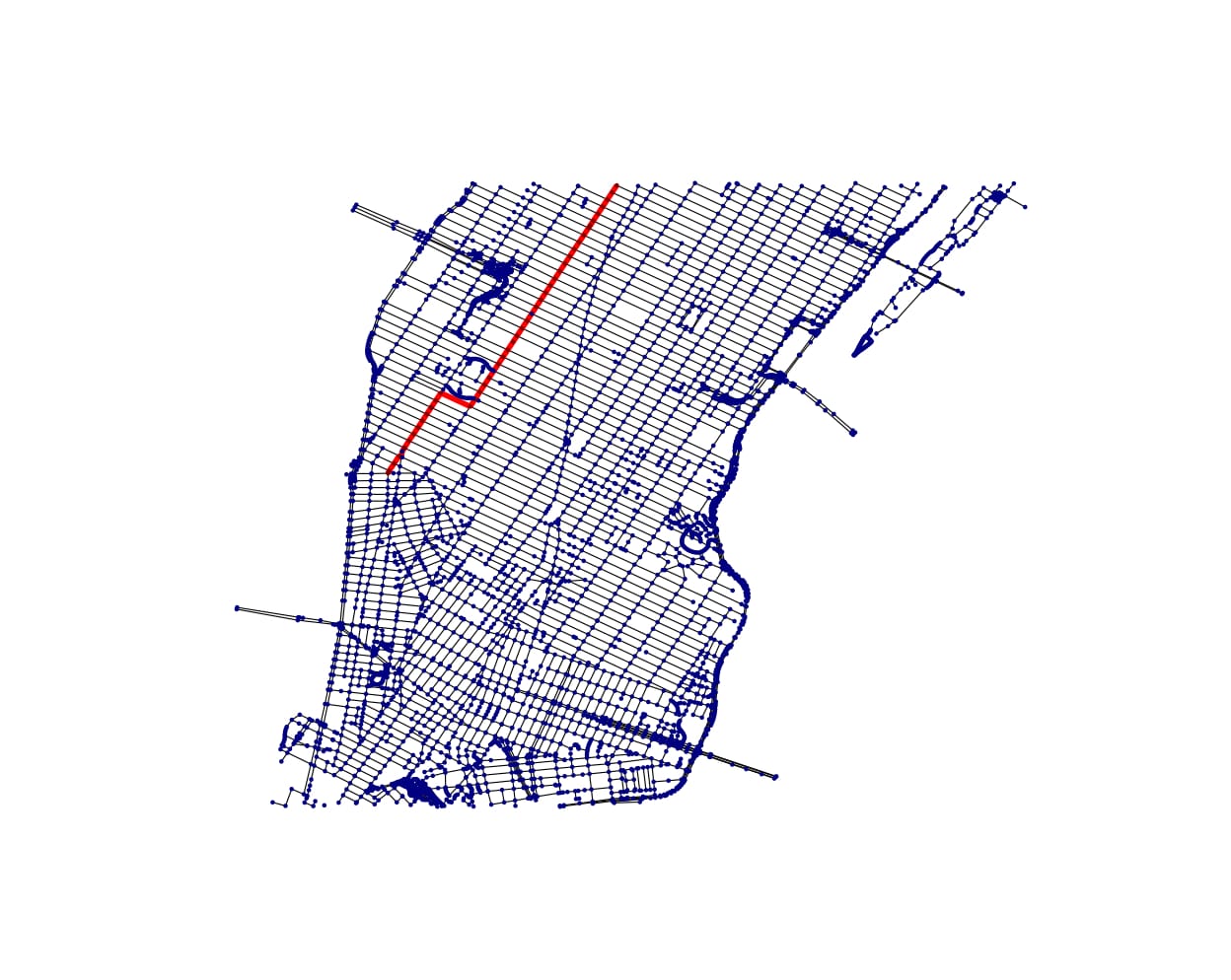}
    \\
    {\small (a) Ground-Truth}  &  {\small (b) GVIN prediction}
    \end{tabular}
  \end{center}
  \caption{\label{fig:nyc_visual} Sample planning trajectories on New York City street map.}
\end{figure*}

\end{document}